\pdfoutput=1
\documentclass[11pt]{article}

\usepackage[final]{acl}

\usepackage{times}
\usepackage{latexsym}
\usepackage[dvipsnames]{xcolor}

\usepackage{amsmath}
\usepackage{subcaption}
\usepackage{graphicx}
\usepackage{float}
\usepackage{dblfloatfix}
\usepackage{caption}
\usepackage[T1]{fontenc}
\usepackage[utf8]{inputenc}

\usepackage[table,xcdraw]{xcolor}
\usepackage{adjustbox}

\usepackage{microtype}

\usepackage{inconsolata}

\usepackage{graphicx}

\usepackage[dvipsnames]{xcolor}
\usepackage{amssymb}% http://ctan.org/pkg/amssymb
\usepackage{pifont}% http://ctan.org/pkg/pifont
\newcommand{\cmark}{\ding{51}}%
\newcommand{\xmark}{\ding{55}}%
\usepackage{tabularx}
\usepackage{booktabs}      % nicer \toprule, \midrule, \bottomrule
\usepackage{array}         % for >{\centering\arraybackslash}
\usepackage{makecell}      % for \setcellgapes
\newcolumntype{Y}{>{\centering\arraybackslash}X}

\title{SteerVLM: Robust Model Control through Lightweight Activation Steering for Vision Language Models}

\author{Anushka Sivakumar \qquad Andrew Zhang \qquad Zaber Hakim \qquad Chris Thomas \\
Department of Computer Science\\
    Virginia Tech \\
    {\tt \{anushkas01,azhang42,zaberhakim,christhomas\}@vt.edu}
}

\begin{document}
\maketitle
\begin{abstract}
This work introduces SteerVLM, a lightweight steering module designed to guide Vision Language Models (VLMs) towards outputs that better adhere to desired instructions. Our approach learns from the latent embeddings of paired prompts encoding target and converse behaviors to dynamically adjust activations connecting the language modality with image context.
% \ct{This clause ",effectively" sounds extremely AI/ChatGPT - try to remove comma' clauses}
This provides fine-grained, inference-time control over complex output semantics without modifying model weights
while preserving performance on off-target tasks. Our steering module requires learning parameters equal to 0.14\% of the original VLM's size.
Additionally, our steering module gains model control via dimension-wise activation modulation and adaptive layer-wise steering without requiring pre-extracted static vectors or manual tuning of intervention points.
Furthermore, we introduce \textsc{Vnia} (Visual Narrative Intent Alignment), a multimodal dataset specifically created to facilitate the development and evaluation of VLM steering techniques. Our method outperforms existing intervention techniques on steering and hallucination mitigation benchmarks for VLMs and proposes a robust solution for multimodal model control through activation engineering. 
% \ct{This last sentence seems very ChatGPT-ish -- would be better to have a sentence summarizing the results (e.g. our method outperforms existing internvention techniques on a number of benchmarks etc"}
\end{abstract}

\section{Introduction}
\begin{figure}[t]
\includegraphics[width=\linewidth]{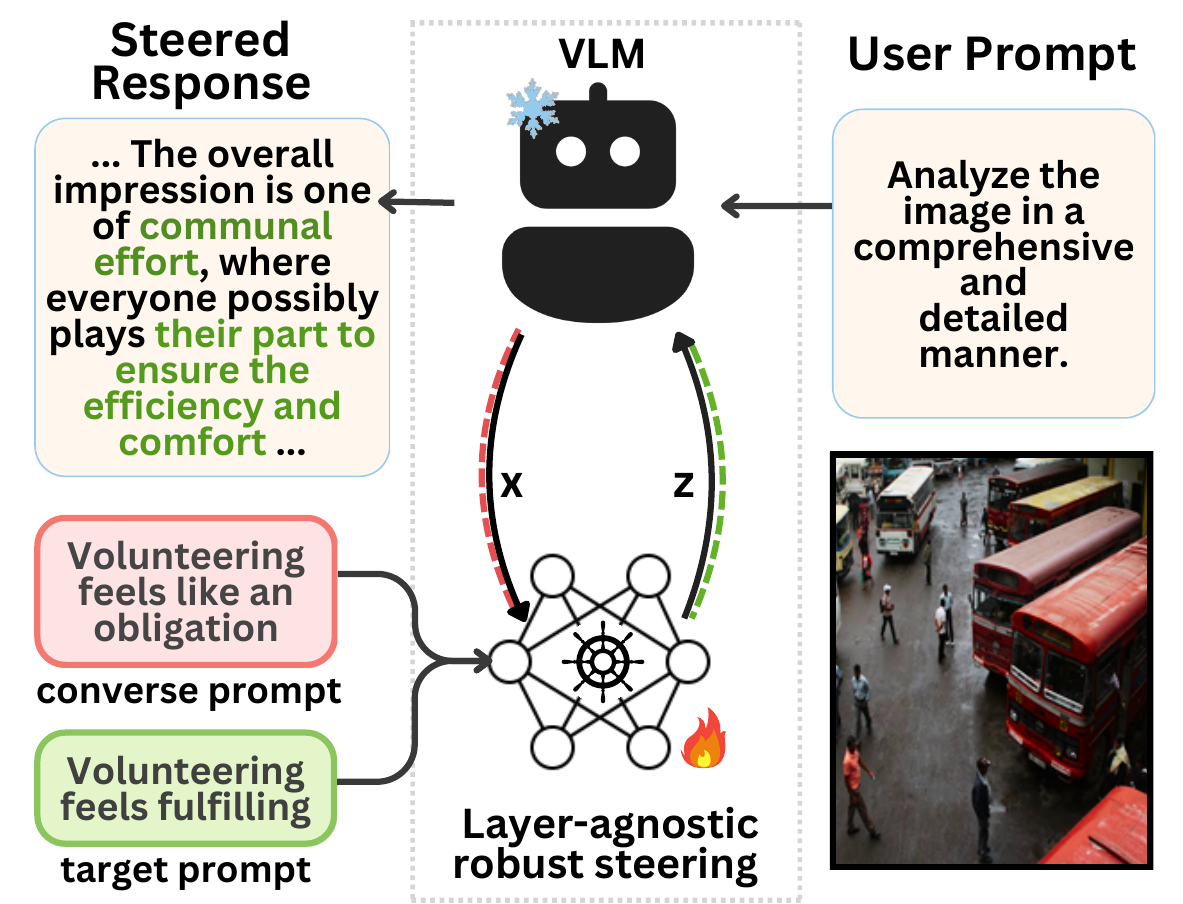}
\centering
\caption{SteerVLM overview. We introduce a layer-agnostic steering module that adjusts the model's output towards a target prompt and away from a converse prompt.
% \ct{the figure is kind of really big and tall -- try to compress to save verical space / people don't like seeing papers that waste space. this could be compressed more. can we also show them what we mean by a prompt / and target / converse prompt and a steered output? if you shrink it vertically you could write what p+ and p- is below to show them, for example? not everyone might understand what you mean by that. also this needs to be on first page}
}
\vspace{-1em}
\label{concept}
\end{figure}
% \ct{In general an intro should follow this structure: introduce context / problem, current solutions for that problem / what those limitations are with existing solutions, how we propose to tackle it, our contributions.}

    Large Vision Language Models (VLMs) demonstrate remarkable capabilities across a wide range of tasks such as image captioning \cite{vinyals2015tellneuralimagecaption}, visual question-answering \cite{VQA}, and more. Yet, effectively eliciting their full potential remains a challenge.
    % \ct{add cites after each of these}\
    % \ct{Extremely wordy -"effectively eliciting their full potential remains a challenge"}
    Naive prompting does not guarantee optimal performance on tasks that are highly dependent on following instructions \cite{Zhou_2022}.
    % \ct{as observed -- kind of weird writing -- optimal performance on what? don't you mean on instruction following - if so need to be explicit}
    Techniques such as Chain-of-Thought prompts \cite{wei2022chainofthought} and few-shot examples \cite{brown2020languagemodelsfewshotlearners} have been shown to significantly improve Large Language Model (LLM) performance without modifying the underlying model. This suggests the presence of an ``elicitation overhang'': a gap between a model's potential and our ability to fully access it \cite{turner2023actadd}. 
    We often lack a complete understanding of how to best extract the desired behavior or knowledge from the model. 
    % \ct{sounds like gpt -- "highlights"}

Prompt engineering is a primary method for guiding model behavior, but its effectiveness 
% \ct{what do you mean by reliability here - don't you mean effectiveness}
diminishes with complex inputs or nuanced desired outputs \cite{ye2022unreliabilityexplanationsfewshotprompting, li2024long}. 
In VLMs, where models must integrate both visual and textual inputs, prompt engineering becomes less effective and makes it harder to elicit good VLM responses.
% \ct{this is really wordy / really AI / bad}
To address the limitations of prompt-based approaches, we present \textbf{SteerVLM}, an inference time intervention based steering approach that aligns model outputs with desired instructions (Figure \ref{concept}).
% To address this elicitation overhang in VLMs, this paper explores an alternative approach gaining traction in the language space: activation engineering \cite{turner2023actadd, ding2024adaptive}. 
% \ct{I wouldn't say we "explore" an alternative approach - this is our method, not an "exploration". I would be more direct: We present an activation engineering approach, though unlike past work, etc...}
Rather than relying solely on input prompts, we utilize latent space vector arithmetic to modify a model's internal activations, specifically its hidden states, during inference. 
% \ct{again -- "investigate" -- it makes it sound like you are publishing a paper like Anthropic that is investigating stuff in claude when actually you are publishing a method}
According to recent literature \cite{turner2023actadd, subramani2024word}, directly modifying the model's activation can more effectively steer a VLM's response than prompt engineering alone. 
% \ct{again sounds gpt ish}

Recent lightweight approaches adjust LLM outputs without resorting to full-scale fine-tuning or prompt-tuning. However, these methods often exhibit several limitations. Existing methods often require extracting static steering vectors from a dataset containing steered and unsteered responses \cite{zhao2024analyzing, rodriguez2024controllinglanguagediffusionmodels, ding2024adaptive}. 
% while others utilize naive contrastive approaches, such as direct activation subtraction, to generate these vectors \cite{turner2023actadd, marks2024steering}. 
Furthermore, many methods require interventions at a predetermined, hyperparameter-tuned layer and apply the same uniform steering vector for each generated token \cite{turner2023actadd, zhao2024analyzing}. Finally, a significant challenge is that they often do not adapt well to the complexities of a multimodal setting \cite{rodriguez2024controllinglanguagediffusionmodels}. All these factors can limit their overall adaptability and effectiveness in steering.
In contrast, we introduce a novel, parameterized steering mechanism that enhances control over model behavior without the inflexibility and potential loss of generalizability associated with fine-tuning. We achieve this by training a lightweight steering module that learns to predict  adjustments by analyzing pairs of prompts that encode both the desired target behavior and its converse. At inference time, this trained module dynamically computes and applies these adjustments to the VLM's intermediate hidden states.  
% Our contributions extend these ideas into the multimodal VLM setting, where language must be grounded in vision.
Our contributions extend these ideas into the multimodal VLM setting, where language must be grounded in vision. Here, the steering is applied to the language module of the VLM - post projection of visual features in the language space. This approach allows us to explore controlled generation in multimodal models, following prior methods that intervene solely on the LLM backbone of VLMs \cite{huang2024opera, liu2024payingattentionimagetrainingfree}.

Unlike prior methods
that are
often limited to direct subtraction of averaged activations \cite{turner2023actadd, marks2024steering}, our module learns a more complex, non-linear mapping from these target-converse activations. This allows it to selectively amplify or suppress relevant activation patterns, effectively learning to ignore irrelevant signals for robust, token-specific steering signals that can be adaptively applied across multiple layers. In  comparison to simple prompting-based techniques, and previous steering techniques \cite{rodriguez2024controllinglanguagediffusionmodels, turner2023actadd, marks2024steering}, SteerVLM is capable of modeling the relationship encoded in semantically rich prompt pairs within the VLM's activation space. This leads to better control over the model's outputs.
% By modeling the nuanced relationships encoded in semantically rich prompt pairs, our steering module captures more intricate patterns within the VLM's activation space, leading to more precise control over its outputs.
% \ct{again it is passive voice like gpt -- I would say Our steering module models the relationships encoded in semantically rich prompt pairs which leads to more precise control over model outputs... something like that}

We also introduce VNIA (Visual Narrative Intent Alignment), a multimodal dataset specifically designed to support the development and evaluation of steering mechanisms for vision language models.
% \ct{do you think it's weird to say a "novel multimodal dataset"}
To our knowledge, VNIA is the first such dataset to provide steered responses directly conditioned on images, addressing a key resource gap for VLM steering research.
% \ct{need to be careful what we mean by topic-aligned textual responses -- i think you mean steered responses, but topic-aligned could mean something else} - makes sense

In summary, our contributions are as follows \footnote{Code and dataset available at \url{https://github.com/22anushka/SteerVLM/}}:
\begin{enumerate}

\setlength\itemsep{1pt}
\item We propose a novel lightweight steering module 
% steering mechanism 
for VLMs 
% that utilizes a lightweight module 
to learn complex, non-linear adjustments from target and converse prompt pairs for finer-grained model intervention.
% , enabling finer-grained control than prior methods.
\item We dynamically apply token-specific steering across multiple layers without predetermined layer selection or static vectors.
% \item We develop an adaptive steering approach where the module dynamically applies token-specific and learns to selectively amplify or suppress relevant activation patterns without requiring pre-extracted static vectors or hyperparameter-tuned layer selection.
\item We present VNIA, the first multimodal dataset providing textual responses conditioned on images and steering directions.

\item We quantitatively and qualitatively evaluate our method against existing steering techniques to show that our approach outperforms previous methods in topic-based steering and off-target task of hallucination mitigation.
% \item \ct{You can say something like We quantitatively and qualitatively evaluate our method against other intervention approaches and show that our approach outperforms existing methods for model intervention and control etc.}
\end{enumerate}

\section{Related Works}
 \label{ch:lit_review}

\subsection{LLM Steering Techniques}
Previous research has introduced numerous steering methods for LLMs. These interventions span weight-based techniques such as supervised fine-tuning \cite{goodfellow2016deep, ouyang2022training, wei2022finetuned}, weight editing \cite{hu2021lora}, and reinforcement learning-based approaches \cite{ouyang2022training, schulman2017proximalpolicyoptimizationalgorithms}. Prompt-level interventions include automated prompt engineering and guided decoding for controlled output or style transfer. Token embedding interventions, like soft prompting, append learnable tensors optimized for specific datasets \cite{lester-etal-2021-power, li-liang-2021-prefix}. Activation-based interventions utilize steering vectors, adjusting activations directly to produce targeted behaviors.

\subsection{Steering Vectors}
Steering vectors can be computed via methods such as mean activation shifts between sentiment prompts \cite{turner2023actadd}, correlating feature labels with attention head activations \cite{li2023inference}, differences in hidden or embedding space vectors \cite{subramani2024word, marks2024steering}, or classifier-based decision boundaries \cite{ding2024adaptive}. These vectors effectively direct activations toward desired outcomes.

\subsection{Activation Engineering and Inference-time Intervention}
Activation engineering facilitates efficient inference time control without full model fine-tuning \cite{zhang2025controllinglargelanguagemodels}. Approaches include latent steering vectors \cite{hernandez2022extracting}, contrastive activation addition \cite{marks2024steering}, and Adaptive Activation Steering (AAS), which dynamically adjusts activations to enhance truthfulness \cite{ding2024adaptive}. Techniques like Concept Eraser \cite{gandikota2025erasingconceptualknowledgelanguage}, concept activation vectors \cite{zhang2025controllinglargelanguagemodels}, activation transport \cite{rodriguez2024controllinglanguagediffusionmodels}, style-specific neurons \cite{hu2024stylespecific}, conceptors \cite{belrose2024steeringconceptors}, and multi-attribute steering \cite{akyurek2025multiattribute} provide diverse, interpretable tools for model control.

Although activation steering has shown efficacy in style transfer, toxicity mitigation, and hallucination reduction, current methods predominantly focus on language alone, restricting their effectiveness to knowledge embedded within textual modalities. Our approach introduces image contexts as an additional modality, requiring the model to produce coherent, contextually relevant responses. Existing methods often struggle with zero-shot transferability and require extensive tuning to identify optimal steering layers and steering vectors across different model architectures (e.g., Llama \cite{touvron2023llama2openfoundation} vs. Qwen \cite{qwen2025qwen25technicalreport}).

Our approach overcomes these limitations by using prompt-embedded activation vectors as steering signals and a layer-agnostic method to identify optimal steering layers, enabling robust and context-aware steering within vision language models.

% \section{Background}
% Write about activation engineering and steering, write about mechanistic interpretability and feature encoding within an llm, talk about vision language models maybe? 

% Can add a lot for background in thesis, talking about the llava paper, its strengths, weaknesses, the diagram from llava, etc. under VLMs in general or whatever

\section{Approach}
\subsection{Steering Module}

    \begin{figure}[t]
\includegraphics[width=\linewidth]{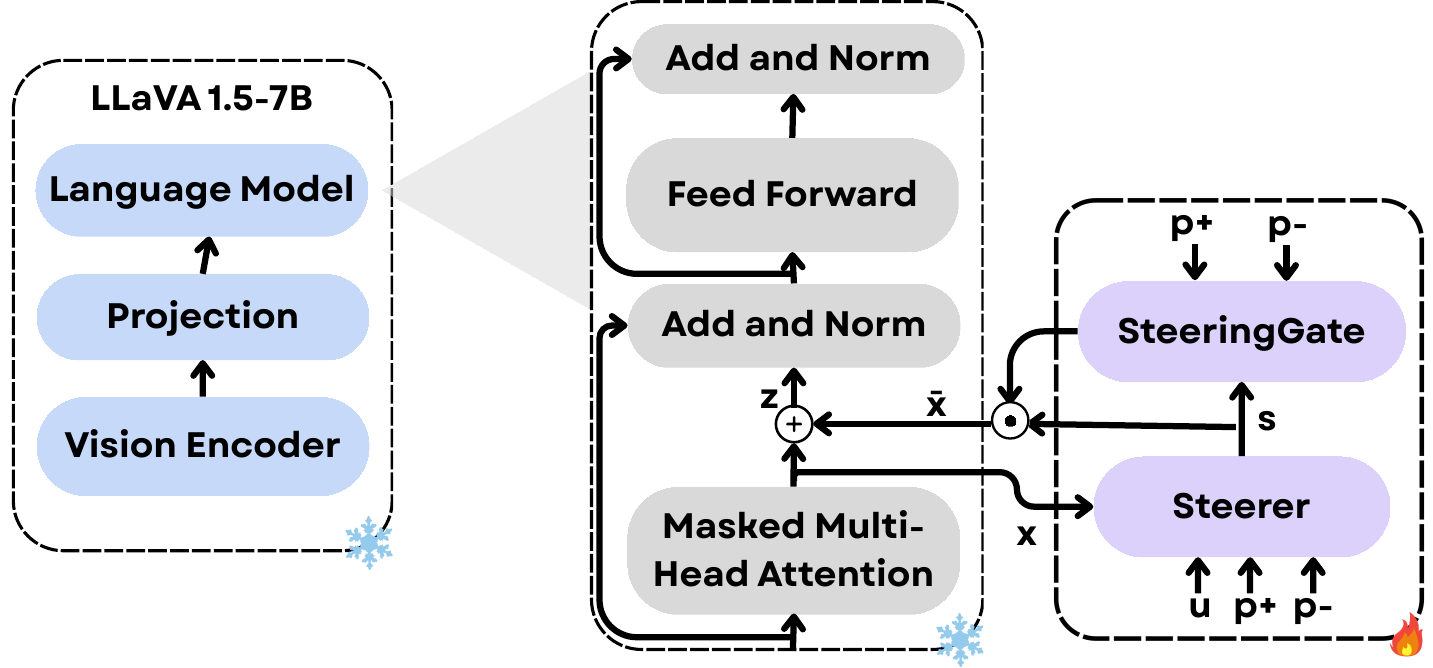}
\centering
\caption{The Steering Module. The steering module is hooked right after the multi-head attention module in each layer of the language decoder. The Steering module consists of the Steerer and the SteeringGate which steer the activations based on the context vectors. The steered activation is added to the residual.}
\vspace{-10pt}
\label{steeringmodule}
\end{figure}
    The Steerer and the SteeringGate form the proposed steering module. 
    % The module connects directly 
    % % the language model within the VLM
    % % , specifically 
    % after the multi-head attention mechanism in each decoder layer of the language model of the VLM (Equation \ref{eqn_1}).
    % The steering module is attached right after   
    % % the language model within the VLM
    % % , specifically 
    % the multi-head attention mechanism in each decoder layer of the language model of the VLM. 
    As shown in Figure \ref{steeringmodule}, activations from the current layer (post multi-head attention) pass through the steering module and are added to the residual stream pre-normalization and before entering the feed-forward layer of the language decoder for that layer. The modified activation is denoted as $z$ (Equation \ref{eqn_1}). With this approach, the steering module learns only the necessary delta to align the activation with the target behavior, while preserving the information already present in the original activation $x$.

    Importantly, the same steering module is shared across multiple layers. 
    This way, the steering module does not need predetermined layers to steer on but rather learns to steer at each individual layer during training.

     Unlike previous methods, this architecture does not depend on pre-extracted steering vectors computed through probing mechanisms on a set of samples representing specific concepts or target behaviors. Instead, it utilizes the activations of the context token from a pair of prompts $(p_{+}, p_{-})$, which denote the target and converse behaviors, respectively. Steering is then performed using these prompt embeddings.
    % Add equation for the attention steerer mechanism here.
    \begin{equation}
    \begin{gathered}
    z_l = x_l + \lambda \bar{x}_l \\
    \bar{x}_l=f(x_l, p_{+_l}, p_{-_l})
    \label{eqn_1}
    \end{gathered}
    \end{equation}
    \begin{equation}
f(x_l, p_{+_l}, p_{-_l}) = g(h(x_l, c_l), p_{+_l}, p_{-_l}) \odot h(x_l, c_l)
    \label{eqn_2}
    \end{equation}

      $x_l$ denotes the activation after the attention operation at layer $l$ before the add and normalize operation in the decoder. $f(x_l, p_{+_l}, p_{-_l})$ denotes the steering module operation in the activation space on $x_l$ with the activations of the target prompt $p_{+_l}$ and converse prompt $p_{-_l}$ at a particular layer $l$. Steering module operation is further broken down in Equation \ref{eqn_2} 
    % $f(x_l, p_+, p_-)$ (Equation (\ref{eqn_2})) is further broken down as $g(h(x_l, c))$
    where $c_l$ denotes context vectors ($\{u_l, p_{+_l}, p_{-_l}\}$ as seen in Figure \ref{steeringmodule} and defined in Section \ref{steerer_section}), $g(.)$ denotes the SteeringGate, and $h(.)$ denotes the Steerer. 
    Additionally, the steering strength $\lambda$ can be adjusted during inference to control the amount of steering applied to the model's activations during the forward pass.

    % Include steering formula and explanation of notation here.
    
    \paragraph{Features Represented by Dimensions}
    Drawing upon observations from research in mechanistic interpretability, we note that features within a layer's activation space are often represented as superpositions of dimensions (i.e., combinations of neurons) \cite{goldengateclaude}. Building on this, our approach focuses on steering across these dimensions. We hypothesize that the features captured from the embeddings of the target and converse prompts can be manipulated along these dimensions to influence the overall feature representation, thereby steering the model's output \cite{cross-diffing}.

\
    The steering module intervenes in the model's latent space to align the generated output with the desired behavior.
    The Steerer determines the necessary adjustments by amplifying or suppressing dimensions to align with the target behavior, while the SteeringGate modulates the amount of steering required per dimension, based on the target behavior and the output from the Steerer.

% Consequently, the SteeringGate can be understood as modulating the degree of steering applied to each dimension by analyzing the steered output in relation to the target and converse prompt pair.

    \subsection{Steerer} \label{steerer_section}
The Steerer is designed to interpret the relationships among the target prompt, the converse prompt, and the current activation. It uses a lightweight, two-layer, multi-head attention architecture to effectively capture subtle interactions within the activations. 

% Insert block diagram of the Steerer here.

% \begin{figure}[htbp]
%     \centering
%     \begin{subfigure}[t]{0.48\linewidth}
%         \centering
%         \includegraphics[width=0.6\linewidth]{figures/steererBlock.pdf}
%         \caption{Steerer}
%         \label{steerer}
%     \end{subfigure}
%     \hfill
%     \begin{subfigure}[t]{0.48\linewidth}
%         \centering
%         \includegraphics[width=0.6\linewidth]{figures/SGblock.pdf}
%         \caption{Steering Gate}
%         \label{steeringgate}
%     \end{subfigure}
%     \caption{Steering Module Block Diagram \ct{should this figure be merged with figure 2 in a two-col figure in some way? it seems very unpolished though. ppeople don't usually like seeing flow-chart type figs -- seems kind of rough -- it can be illustrated better}} - would it be ok to let this be for now. I dont know how to neatly integrate it into figure 2 and as of now I like figure 2 so I might want to wait before I think of making changes
%     \label{steering_module_arch}
% \end{figure}

The Steerer's architecture starts with a down-projection layer that reduces the model's dimension to one-eighth of its original size. This reduction significantly decreases the parameter count, making the model lightweight.
% Finally, an up-projection layer restores the dimensionality to match the original embedding size.

The Steerer receives four concatenated activations\footnote{":" denotes concatenation along the sequence dimension} as input: the current activation from the main model $x$, the unsteered activation $u$ from the current layer, and the target-converse activation $p_+$ and $p_-$. $u$ is the activations for the same set of inputs to the model but without any steering on prior layers. This helps provide context about cumulative steering effects from earlier layers. $\{u, p_{+}, p_{-}\}$ are considered as the context vectors $c$.

% Add equation for input to the Steerer here.
\vspace{-20pt}
\begin{equation}
\begin{gathered}
c = u':p'_{+}:p'_{-},\\
j(x,c) = MHA_{1}\bigl(q = x';\;k,v = x':c\bigr),\\
s=h(x,c)= W_{up}\bigl(MHA_{2}(q = j;\;k,v = j:c)\bigr)
\end{gathered}
\label{eqn:three}
\end{equation}

where the Steerer function is denoted by $h(.)$, and the superscript $'$ denotes down projected input vectors from $d\_model$ (embedding dimension of the VLM) to a lower dimension and $W_{up}$ denotes the up projection of the dimensions back to $d\_model$.

The model utilizes a combination of self-attention and cross-attention where $x$ serves as the query and $x$, $u$, $p_+$, and $p_-$ concatenated together serve as the keys and values.
The steerer works along the dimension of the tokens to capture complex relationships between the input vectors. This approach, in comparison to a contrastive approach, is effective even when the prompt pair does not consist of exact opposites. For example "Volunteering feels fulfilling" versus "Volunteering feels obligatory,". Fulfillment and obligation, while not direct antonyms, represent mutually exclusive sentiments in the given context. The attention mechanism captures such intricacies in semantically rich prompt pairs.

    \begin{figure}[H]
\includegraphics[width=0.9\linewidth]{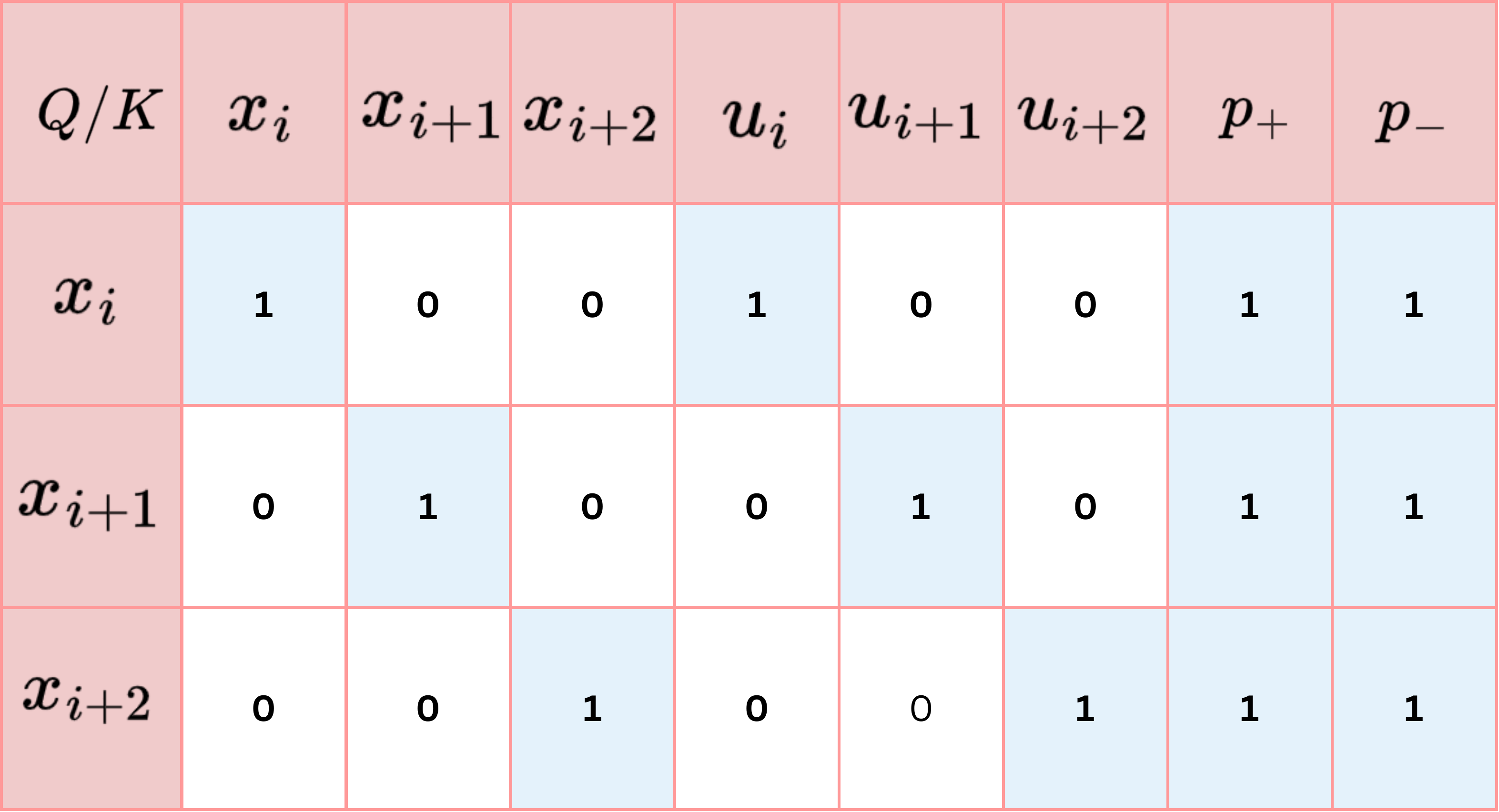}
\centering 
\caption {Attention mask for the Steerer's Attention Block. $i$ denotes token at timestep 0 and $i+1$ denotes token at timestep 1. We make use of a boolean mask here where 1, 0 denote unmasked tokens and masked tokens respectively. }
\label{attn_mask}
\vspace{-1.0em}
\end{figure}

\paragraph{Attention Mask} The attention module within the steerer employs a sparse attention mask. Each token $x_i$ attends only to: itself $x_i$, its unsteered counterpart $u_i$, the target prompt token $p_+$, and the converse prompt token $p_-$ as seen in Figure \ref{attn_mask}. The attention module within the VLM already computes attention over previous tokens, so the Steerer uses this focused cross-attention to choose the right steering pattern for each token.
% By focusing on these specific relationships, this approach minimizes attention dispersion to potential distractors within the sequence. 
% \ct{this part of the text was unclear to me -- I'm sure reviewers won't understand it eithe r-- we should try to clarify more}

    \subsection{SteeringGate}

    The SteeringGate is a multilayer perceptron (MLP) designed to regulate the amount of steering applied at each dimension. It determines steering intensity based on relationships among the target prompt, the converse prompt, and the steered activations. The SteeringGate takes in $s, p_+, p_-$, as inputs.

Similar to the Steerer, the SteeringGate uses down-projection and up-projection layers, ensuring a lightweight yet effective structure.

The MLP captures complex, nonlinear relationships among the computed steering vector and the pair of target-converse prompts. A sigmoid gating mechanism is applied to the output for \textbf{dimension-specific} control over the steering intensity. This is formulated as,

\vspace{-20pt}
\begin{equation}
g(s, p_{+}, p_{-}) = \sigma\left(W_{up}\left(MLP(s',p'_{+},p'_{-})\right)\right)
\end{equation}

where \( g(.) \) is the SteeringGate function, and \( {W_{up}} \) is the up-projection layer, restoring the dimensions to match $d\_model$. The superscript \( '\,\) indicates down-projected input vectors. The inputs \( s', p'_{+}, p'_{-} \) are concatenated along the hidden dimension before being fed into the MLP.
% \ct{can't Up or Down be like W\_up, W\_down, etc} -- dont want to take up more space with the W\_down

% Because the SteeringGate's primary role is adjusting steering intensity across different dimensions per layer, reducing dimensionality helps it capture broader relationships among activations. This design enables the SteeringGate to identify when similar features vary in strength across dimensions, thereby effectively prioritizing dimensions requiring stronger steering adjustments.

\section{Dataset}
\label{dataset}

We also contribute \textbf{VNIA (Visual Narrative Intent Alignment), a multimodal steering dataset} designed to train and evaluate our steering module. We randomly sampled 61,391 images from the CC3M dataset \cite{sharma-etal-2018-conceptual} and generated steered responses to prompts using Qwen2.5-VL-72B \cite{qwen2025qwen25technicalreport} (see Appendix for details). Figure \ref{dataset_fig} illustrates the complete process used to generate this dataset.

\begin{figure}[H]
\includegraphics[width=0.8\linewidth]{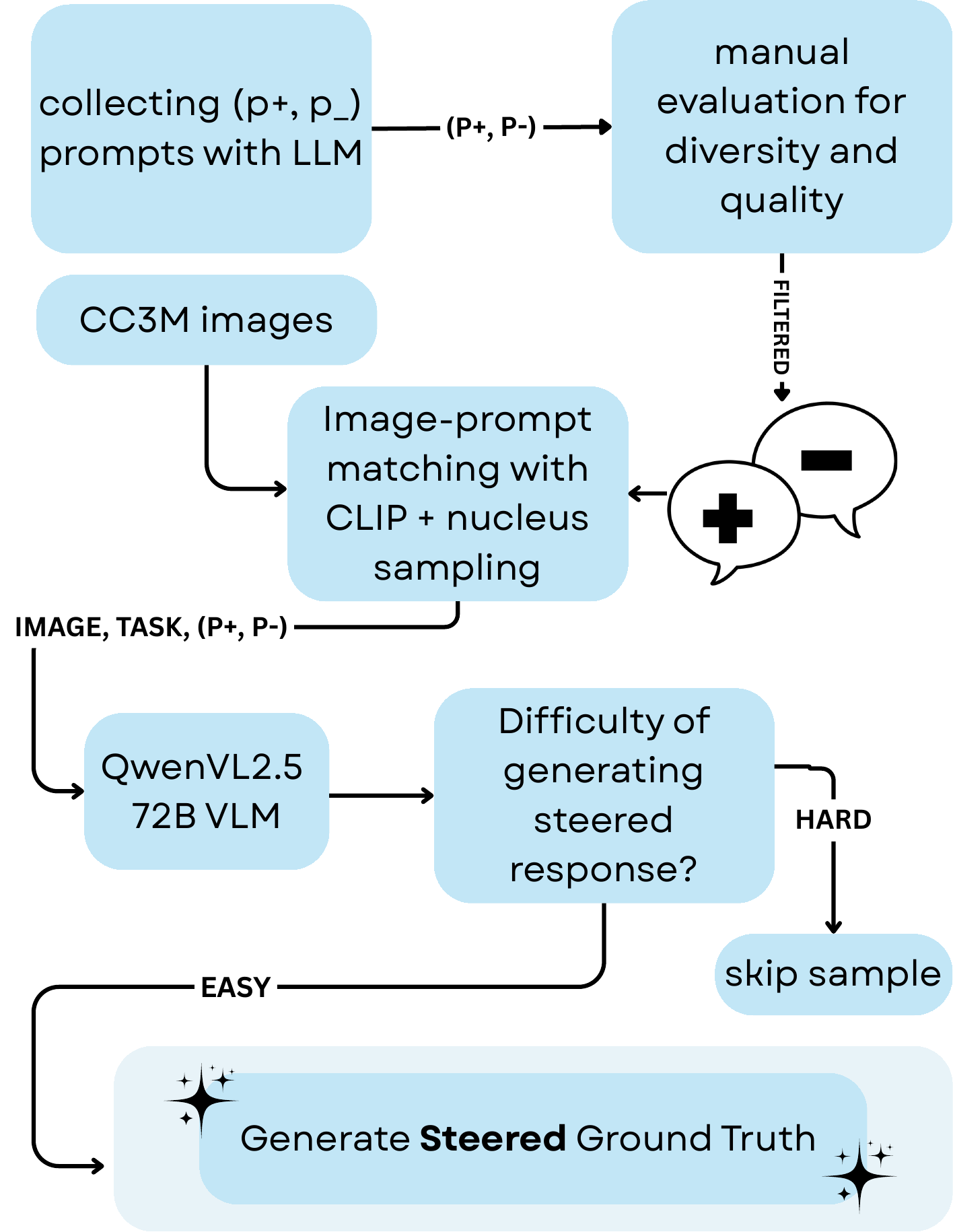}
\centering 
\caption{VNIA Dataset synthesis pipeline. We begin by generating target/converse prompt pairs. The prompts are then paired with images using CLIP-score matching with adaptive nucleus sampling for diversity. Finally, steered and unsteered responses are generated by Qwen2.5-VL-72B VLM.}  
% \ct{I wouldn't call it a flowchart... try to make more graphical like in the JourneyBench dataset figure. Also move closer to the portion of text where it is discussed. Figs should be on the same page as refers to them when possible}

\label{dataset_fig}
\end{figure}

\paragraph {Prompt Sampling} First, we used GPT-4o \cite{openai2024gpt4ocard} to generate pairs of mutually exclusive prompts for various topics (see Appendix Table \ref{tab:steering_prompt_gen}). These prompts were then manually filtered and modified to align with our steering objectives. Include a wide variety of topics that explore emotional states, daily activities, and abstract themes. The goal is to create a list with contrasting pairs that span these areas, offering a rich mix of relationships between positive and negative perspectives.

The pairs of target and converse prompts were manually inspected by the authors. Since the dataset contains only 463 prompts, applying human-in-the-loop techniques with a set of filtering criteria was manageable. The criteria included ensuring diversity in both the topics of the prompts (e.g., cooking, running, astronomy, etc.) and their semantics (e.g., love–hate, easy–difficult, intriguing–confusing), removing duplicates, and ensuring mutually exclusive semantics. For example, consider the pair "Volunteering is fulfilling" vs. "Volunteering is time-consuming." While both can be true simultaneously, similar to the approach in \cite{turner2023actadd}, we aimed to ensure a degree of mutual exclusivity between the target and converse prompts.

Next, the set of prompts was split into distinct training and evaluation sets. We then matched these prompts with the sampled images using CLIP embedding scores.

\begin{figure}[t]
\includegraphics[width=0.9\linewidth]{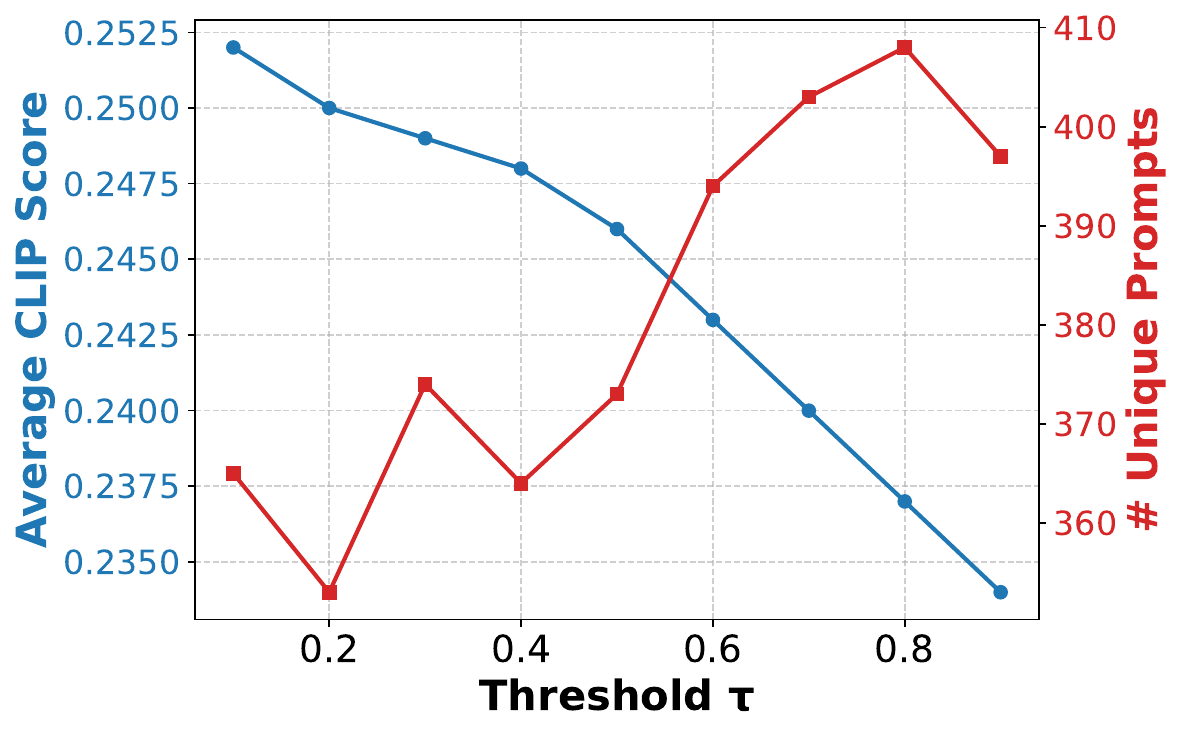}
\centering 
\caption{Entropy threshold analysis to analyze trade-off between diversity and matching between image and prompt pairs.
}
\label{entropy_threshold}
\vspace{-0.5em}
\end{figure}

\paragraph {Image-Prompt Pairing} A key requirement was to ensure the steering prompts were relevant to the image content but not so trivially descriptive that the steering task offered little challenge. To achieve this balance between relevance and difficulty when selecting the final image-prompt pairs, we employ an adaptive, entropy-based nucleus sampling threshold where the value of top$-p$ is selected based on the sharpness or flatness of the probability distribution obtained from softmax-normalized CLIP outputs. This method ensures that the chosen image and steering prompt  pair share some correlation without being obviously or directly linked. As seen in Figure \ref{entropy_threshold}, we conduct an ablation study for entropy thresholds $\tau \in [0.1, 0.9]$ (over a batch size of $1024$) to quantify the trade-off between CLIP scores and diversity measured by the number of unique prompts. In qualitative terms, overly low thresholds select a small subset of the target versus converse prompts lowering the diversity of the prompts in the dataset whereas overly high thresholds increase sample rejection and random prompt assignments, undermining dataset consistency and size. By adopting 
 $\tau = 0.6$ as our default (with an optional 
 $\tau= 0.7$ setting for diversity-critical applications), we ensure an optimal balance between steerable outputs and steering signals.

\paragraph{Generating Steered Responses} During the dataset generation phase, we prompted Qwen2.5-VL-72B to assess the difficulty of producing a steered output for a given image and prompt pair (target and converse). Samples deemed too difficult were pruned.
% Samples deemed too easy or too difficult were pruned.
% accordingly. 
We constructed two forms of steering prompts for the dataset: one based on descriptive image captions and the other based on creative short stories. The image description prompts are adapted from those proposed in the LLaVA paper \cite{liu2023visualinstructiontuning, liu2024improvedbaselinesvisualinstruction} and include variations requesting both short, concise answers and detailed, descriptive responses.

We utilize the generated VNIA dataset throughout our training process, including supervised fine-tuning (SFT), as well as during the evaluation stage to assess the quality of the steered generation. Detailed descriptions of the prompts and examples of the steering pairs can be found in the Appendix.

 \section{Training}
% Initially, we construct a multimodal steering training and validation dataset as detailed in Section \ref{dataset}. 
The supervised fine-tuning stage is essential to stabilize the steering module due to the random initialization of model weights and the shared nature of the architecture across layers. The SFT alignment stage provides quick adaptation to the steering task.
 
Given that our dataset (Section \ref{dataset}) size is approximately 20-25\% smaller than typical supervised fine-tuning datasets and considering stabilization requirements of the shared module across all layers, we trained SteerVLM on 8 A100 GPUs for 5 epochs. Additionally, we keep steering strength $\lambda$ as 1, and learning rate as $3e-4$ with cosine learning rate scheduler during training.

We used standard cross-entropy loss to optimize log probabilities of tokens aligned towards the target behavior.
% \ct{you could throw in a sentence about we also experimented with GRPO and refer them to our appendix}

% $$\mathcal{L}(\mathcal{U}) = \sum_{i=1}^{N} \log \left( \mathbb{P}(u_i \mid u_{i-k}, \ldots, u_{i-1}, \Theta) \right)
% $$

\section{Evaluation Setup} \label{evaluation_setup}
%%% Renamed for clarity - covers data and initial setup
We compare our approach against prior works using an evaluation set derived from the \textsc{Vnia} benchmark. We extract five topic steering vectors from \textsc{Vnia} and generate 150 image-prompt pairs per vector as training data for methods requiring examples to construct steering vectors. An additional set of 20 samples per vector is held out for evaluation. We extract steering vectors for methods that require it from the training set. \textit{For all experiments, SteerVLM is applied in a zero-shot setting, i.e.~on unseen $p_+/p_-$.} GPT-o4-mini \cite{openai2025o4mini} is used as the judge model to evaluate the performance on this task (Appendix \ref{judge_prompt_subsection}).

We also benchmark our work on hallucination mitigation. We use 1,000 training samples from the \textsc{OHD} dataset \cite{liu2024investigatingmitigatingobjecthallucinations}, which consists of \textsc{COCO} \cite{lin2015microsoftcococommonobjects} images paired with both faithful and hallucinated captions. For baselines requiring negative examples to compute steering vectors, we randomly select one hallucinated caption from the “adversarial” category for each training instance. Evaluation is conducted using the standard \textsc{OHD} benchmark, which reports POPE $F_1$ score and accuracy across the popular, adversarial, and random categories \cite{li2023evaluatingobjecthallucinationlarge}. As before, our method operates zero-shot, using handcrafted positive (target behavior) and negative (converse behavior) prompts related to hallucination (listed in Appendix Table ~\ref{tab:behavior_prompts}).
%%% Needs Appendix reference label

\section{Experiments}
\label{experiments}

\begin{table*}[ht]
  \centering
  \small
  \setlength{\tabcolsep}{4pt}    % tighten horizontal padding

  % scale to exactly 75% of full text width and center
  \begin{adjustbox}{width=0.75\textwidth,center}
    \begin{tabular}{|l|c|c|c|c|c|}
      \hline
      \hline
      \multicolumn{1}{|c|}{\textbf{Model}} &  \multicolumn{1}{|c|}{\textbf{Zero-shot}}
        & \multicolumn{4}{c|}{\textbf{Accuracy / F1-Score}} \\ \hline
        &
        & \textbf{Adversarial} 
        & \textbf{Popular} 
        & \textbf{Random} 
        & \textbf{Overall} \\ 
      \hline\hline

      LLaVA1.5-7B Zero-shot \cite{liu2024improvedbaselinesvisualinstruction}
        & \cmark 
        & 79.8/81.7 
        & 85.5/86.1 
        & 88.3/88.8 
        & 84.5/85.5 \\ 
      \hline

      ML-ACT \cite{rodriguez2024controllinglanguagediffusionmodels}
        & \xmark
        & 79.8/75.7 
        & 80.8/76.6 
        & 81.1/76.8 
        & 80.6/76.4 \\ 
      \hline

      MLLM Steering \cite{zhao2024analyzing}
        & \xmark
        & 72.4/76.1 
        & 76.9/79.1 
        & 78.4/80.5 
        & 75.9/78.6 \\ 
      \hline

      CAA \cite{marks2024steering}
        & \xmark 
        & 53.0/68.0 
        & 54.9/68.9 
        & 60.6/71.6 
        & 56.2/69.5 \\ 
      \hline

      Contrastive / layer 
        & \cmark
        & 79.8/81.8 
        & 85.7/86.4 
        & 89.3/89.4 
        & 84.9/85.7 \\ 
      \hline

      Act Add  \cite{turner2023actadd}
        & \cmark
        & 79.2/81.3 
        & 85.7/86.3 
        & 89.3/89.5 
        & 84.7/85.7 \\ 
      \hline

      ACT \cite{ding2024adaptive}
      & \xmark
      & 79.0/80.7
      & 85.5/85.9
      & 89.0/88.9
      & 84.5/85.1 \\
      \hline

      \textbf{Ours}
        &\cmark
        & \textbf{81.5/82.5} 
        & \textbf{87.6/87.7} 
        & \textbf{90.2/90.1} 
        & \textbf{86.4/86.8} \\ 
      \hline
    \end{tabular}
  \end{adjustbox}

  \caption{Evaluation on the OHD \cite{liu2024investigatingmitigatingobjecthallucinations} dataset on the POPE metric. The best scores are highlighted in bold. Zero-shot implies there was no precomputed steering vectors on any types of hallucination mitigation datasets.}
  \label{pope}
\end{table*}

\begin{table*}[ht]
  \centering
  \small
  \setlength{\tabcolsep}{4pt}    % tighten horizontal padding

  % scale to exactly 75% of full text width and center
  \begin{adjustbox}{width=0.75\textwidth,center}
    \begin{tabular}{|l|c|c|c|c|c|c|}
      \hline
      \multicolumn{1}{|c|}{\textbf{Model}}
        & \textbf{sv1}
        & \textbf{sv2}
        & \textbf{sv3}
        & \textbf{sv4}
        & \textbf{sv5}
        & \textbf{Overall}\\ \hline\hline

      ML-ACT \cite{rodriguez2024controllinglanguagediffusionmodels}
        & 0.46 & 0.475 & 0.485  & 0.49  & 0.44   & 0.47\\ \hline

      MLLM Steering \cite{zhao2024analyzing}
        & 0.49 & 0.56 & 0.51 & 0.485 & 0.535   & 0.51 \\ \hline

      CAA \cite{marks2024steering}
        & 0.55 & 0.65  & 0.61  & 0.47  & 0.57   & 0.57 \\ \hline

      Contrastive / layer
        & 0.53 & 0.58  & 0.55  & 0.50  & 0.56   & 0.54\\ \hline

      Act Add \cite{turner2023actadd}
        & 0.52 & 0.60  & 0.59 & 0.475  & 0.58   & 0.55 \\ \hline

    ACT \cite{ding2024adaptive}
        & 0.56 & 0.54  & 0.55 & 0.535  & 0.59   & 0.55 \\ \hline

      \textbf{Ours}
        & \textbf{0.84}
        & \textbf{0.69}
        & \textbf{0.83}
        & \textbf{0.56}
        & \textbf{0.63}
        & \textbf{0.71}
        \\ \hline
    \end{tabular}
  \end{adjustbox}

  \caption{Topic-steering evaluation for 5 steering vectors, evaluated by the judge model. The scores represent an average on a scale of 0-1. The best scores are highlighted in bold.}
  \label{topic_steering}
\end{table*}

\begin{table}[ht]
  \centering
  \resizebox{\columnwidth}{!}{%
    \begin{tabular}{|l|c|c|c|c|}
      \hline
      \textbf{Model}      & \textbf{C1} & \textbf{C2} & \textbf{C3} & \textbf{C4} \\ \hline \hline
      ML-ACT      \cite{rodriguez2024controllinglanguagediffusionmodels}        & \cmark                  & \cmark                 & \cmark                    & \xmark                       \\ \hline
      MLLM Steering \cite{zhao2024analyzing}      & \xmark                  & \xmark                 & \xmark                    & \xmark                       \\ \hline
      CAA       \cite{marks2024steering}          & \xmark                  & \xmark                 & \xmark                    & \xmark                       \\ \hline
      Act-Add       \cite{turner2023actadd}      & \xmark                  & \xmark                 & \xmark                    & \cmark                       \\ \hline
      ACT       \cite{ding2024adaptive}      & \cmark                  & \cmark                 & \xmark                    & \xmark                       \\ \hline
      Ours                & \cmark                  & \cmark                 & \cmark                    & \cmark                       \\ \hline
    \end{tabular}%
  }
  \caption{Comparison of properties of steering methods where C1 denotes \textit{Layer Agnosticity}, C2 denotes \textit{parameterized}, C3 denotes \textit{Dynamic Steering}, and C4 denotes \textit{Zero-shot Steering}.}
  \label{tab:property_comparison}
    \vspace{-1.5em}
\end{table}

% \ct{I think due to the wordiness, we are somehow missing the clear message we want to convey -- that we do not train on certain p / n prompts, yet include these hallucination prompts in our test set, and we show that our model works better on the, despite having not seen them. }

% \ct{The language in this section makes the messages you're trying to get across very muddy. can we prompt gpt to preserve cites but write with active voice, non-wordy style like cvpr paper? this language is losing the contribution / focus. i think we want to make the point that, unlike the methods we compare against, we operate in a zero-shot manner (maybe even include a zero-shot? column in the table with checks or x (see Hani's paper).} - I added a separate table with it.

We compared our proposed method against several baseline and state-of-the-art steering techniques adapted to the LLaVA architecture: ActAdd \cite{turner2023actadd}, ML-ACT \cite{rodriguez2024controllinglanguagediffusionmodels}, CAA \cite{marks2024steering}, ACT \cite{ding2024adaptive} and \cite{zhao2024analyzing} which are primarily designed for steering. Additionally, we included a baseline involving contrasting the activations of the prompt vectors $(p_+, p_-)$ at each layer of the language model.
All experiments were run with temperature = 0.6 and top-$p$ = 0.9 setting \cite{liu2024improvedbaselinesvisualinstruction}.
%%% The following paragraphs describe results and belong ideally in a dedicated "Results" section or subsection. I have improved the writing here but recommend moving this content.

Table \ref{pope} summarizes our results on the effects steering on hallucination mitigation. We make the following observations.
\textbf{SteerVLM achieves state-of-the-art results in zero-shot hallucination mitigation on the OHD dataset \cite{liu2024investigatingmitigatingobjecthallucinations} benchmark.}
Our method achieves superior performance in a zero-shot setting, unlike other methods that extract steering vectors from hallucination mitigation dataset. Notably, SteerVLM improves overall accuracy by 1.7\% and $F1$ score by 0.9\% over \cite{turner2023actadd}. This suggests that carefully designed prompts representing both target and contrasting behaviors can effectively direct activations toward more truthful outputs. This conclusion is further supported by the performance of ActAdd \cite{turner2023actadd} and the baseline Contrastive method, both of which relied on randomly sampled positive and negative prompts (defined in Appendix Table \ref{tab:behavior_prompts}) during their respective processes. 
\begin{table}[ht]
  \centering
  \resizebox{\columnwidth}{!}{%
    \begin{tabular}{|l|c|c|c|}
      \hline
      \textbf{Model}      & \textbf{sv2} & \textbf{sv5} & \textbf{Overall} \\ \hline \hline
      ML-ACT      \cite{rodriguez2024controllinglanguagediffusionmodels}        & 2.4 & 1 & 1.7                      \\ \hline
      MLLM Steering \cite{zhao2024analyzing}      & 3.8 & 3.8 & 3.8 \\ \hline
      CAA       \cite{marks2024steering}          &       6.4 &   6 & 6.2               \\ \hline
      Act-Add       \cite{turner2023actadd}      & 5 & 3.8 & 4.4 \\ \hline
      ACT       \cite{ding2024adaptive}      & 5.2                 & 5                 & 5.1                       \\ \hline
      Ours                & \textbf{8} & \textbf{6.2} & \textbf{7.1}            \\ \hline
    \end{tabular}%
  }
  \caption{Blind Human evaluation on randomly selected examples of steering vector 2 and steering vector 5. Average scores out of 10. The best results are highlighted in bold.}
  \label{tab:human_eval}
  \vspace{-1.5em}
\end{table}

Table \ref{topic_steering} summarizes the performance of steering techniques on the \textsc{Vnia} evaluation dataset.
\textbf{SteerVLM outperforms existing methods in zero-shot steering on the \textsc{Vnia} dataset.}
 Our method surpasses existing intervention techniques and performs 21\% better than the best-performing baseline approach on all steering vector evaluation subsets. While other methods extract steering vectors based on a sample set of the responses steered using $p_+, p_-$ prompts, we operate in a zero-shot manner without having seen the set of $p_+, p_-$ prompts before. ML-ACT \cite{rodriguez2024controllinglanguagediffusionmodels} encountered difficulties building efficient Optimal Transport (OT) maps for each topic on multimodal data. This resulted in NaN values or degenerate answers in some instances, indicating unsuccessful steering in the multimodal setting with its vector-based approach. Similarly, \cite{zhao2024analyzing}'s method also struggled to steer captions effectively, frequently producing empty strings in response to prompts when using its approach based on fine-grained steering vectors derived from a steering prompt rather than a single token of interest.
 
In contrast, our method demonstrated improved quality of steered responses. The experiment was evaluated using the Qwen2.5-VL-72B model \cite{qwen2025qwen25technicalreport} as an automated judge, based on criteria detailed in the Appendix Table \ref{vlm_judge}.
%%% Needs Appendix reference label

Table \ref{tab:human_eval} summarizes the average performance of the model on randomly selected examples of 2 randomly selected steering vectors from the evaluation set of steering vectors. 
\textbf{SteerVLM achieves better qualitative results on blind human evaluation in comparison to existing steering methods.} We made use of the same judge prompt (Appendix Table \ref{vlm_judge}) to evaluate the steered responses.

\section{Ablation Study}

%%% This section describes the setup of the ablation study (methods) and then immediately presents results. Consider separating these.
To justify the architectural choices of our steering module, we conducted ablation studies comparing performance against several variations and baselines. Qwen2.5-VL-72B \cite{qwen2025qwen25technicalreport} is used as the judge model with the judging criteria detailed in the Appendix \ref{judge_prompt_subsection}. 
Table \ref{ablations} %%% Needs Table reference label
presents the ablation results on the \textsc{Vnia} evaluation dataset containing 384 samples.
\begin{table}[H]
\centering
\begin{tabular}{|l|c|}
\hline

% \rowcolor[HTML]{CFE2F3} 
\multicolumn{1}{|c|}{\textbf{Experiment name}} & \textbf{Score} \\ \hline \hline
Zero-shot prompting                                             & 0.57           \\ \hline
One-shot prompting                                              & 0.35           \\ \hline
No steeringGate (SG)                                            & 0              \\ \hline
Same SG sigmoid across dim                                      & 0.59           \\ \hline
No Unsteered activations                                        & 0.69           \\ \hline
\textbf{Ours}                                                   & \textbf{0.78}  \\ \hline
Ours (Specific layers)                                          & {\underline{0.75}}     \\ \hline
\end{tabular}

\caption{Ablation studies justifying architectural and logical choices of the steering module. The best scores are highlighted in bold.}
\label{ablations}
\end{table}

\begin{table*}[ht] \ContinuedFloat
\centering % Centers the entire two-column layout on the page

% ---------- Left Column (image only) -----------------
\begin{minipage}[t]{0.15\textwidth} % 't' for top alignment
  \vspace{0pt} % Ensures the minipage's reference point is at its very top for [t] alignment
  % \centering % Centering the image within the minipage, if needed.
             % If image is \linewidth, centering is implicit.
  \includegraphics[width=\linewidth]{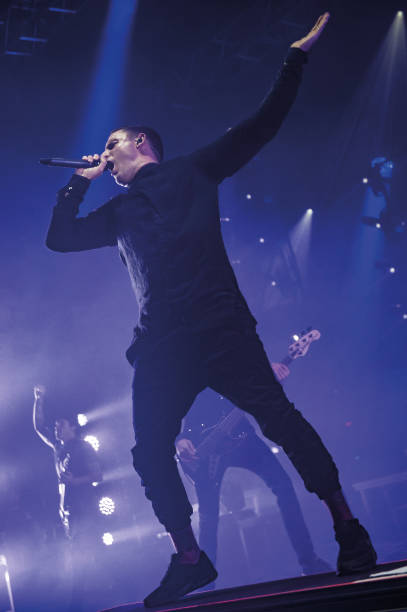} % Image fills this minipage's width
  \\
  \small \textbf{Task:} Describe the image concisely.\par

\end{minipage}% <--- IMPORTANT: % comments out the newline, preventing spurious space
\hspace{\fill}% <--- This provides flexible space between minipages. NO blank line before/after.
% ---------- Right Column (Beliefs/Prompt Text + Comparison Table) -----------------
\begin{minipage}[t]{0.82\textwidth} % 't' for top alignment
  \vspace{0pt} % Ensures the minipage's reference point is at its very top for [t] alignment

  { % Start a group for specific text formatting for beliefs/prompt
    \small % Apply small font size to this text block
    \raggedright % Left-align text. \parindent is 0pt.
    % Adjust parskip for minimal but distinct lines, or set to 0pt if \\ is preferred.
    % For very tight lines, consider removing \parskip adjustment or using \\ with negative spacing.
    \setlength{\parskip}{1pt plus 0.5pt minus 0.5pt} 
    \textbf{Target Prompt:} Learning new skills is overwhelming\par
    \textbf{Converse Prompt:} Learning new skills is exciting\par
    % \vspace{0.2em} % Removed; relying on \parskip or natural line spacing.
    % \textbf{Task:} Describe the image concisely.\par % The \par here will also use the \parskip above.
  } % End group for text formatting
  
  % Adjust space between the belief/prompt text block and the table.
  \vspace{0.3em} % Reduced from \medskipamount for tighter packing. Adjust as needed.

  \begin{tabularx}{\linewidth}{|>{\bfseries}l|X|} % Use \linewidth to fit table within this minipage
  \hline
  SteerVLM &
  \small The individual in the image appears to be on stage, possibly performing ...  The expression on their face is uncertain, possibly conveying a range of \textcolor{LimeGreen}{\textbf{emotions, from focus to concentration. The overall atmosphere seems intense, with the person possibly feeling the weight of the moment.}} The setting and the person's posture may evoke \textcolor{LimeGreen}{\textbf{a sense of pressure or a high stakes situation}}, as they stand on stage. \\ \hline
  Prompting & \small
  The man on stage is wearing a black shirt and has his arms up in the air, possibly \textcolor{BrickRed}{\textbf{expressing excitement or enthusiasm}}. He appears to be the main focus of the scene, possibly indicating .... \\ \hline
  \end{tabularx}
\end{minipage}

% Caption for the entire table* environment
\caption{Steered with SteerVLM vs.\ Prompt-Engineered Task prompt captions produced for the same task and prompt pairs.}
\label{tab:qualitative} % Updated label
\end{table*}

\begin{table*}[ht] \ContinuedFloat
\centering % Centers the entire two-column layout on the page

% ---------- Left Column (image only) -----------------
\begin{minipage}[t]{0.20\textwidth} % 't' for top alignment
  \vspace{0.5em} % Ensures the minipage's reference point is at its very top for [t] alignment
  % \centering % Centering the image within the minipage, if needed.
             % If image is \linewidth, centering is implicit.
  \includegraphics[width=\linewidth]{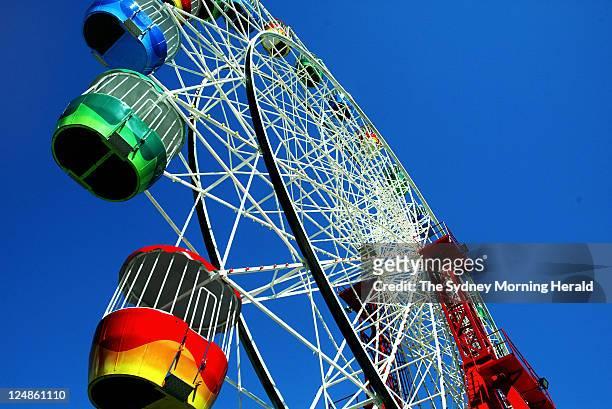} % Image fills this minipage's width
  \\
  \small \textbf{Task:} Write a short story.\par
  \setlength{\parskip}{1pt plus 0.5pt minus 0.5pt}
  \small
    \textbf{Target Prompt:} Bright colors are energizing\par
    \small
    \textbf{Converse Prompt:} Bright colors are overwhelming\par

\end{minipage}% <--- IMPORTANT: % comments out the newline, preventing spurious space
\hspace{\fill}% <--- This provides flexible space between minipages. NO blank line before/after.
% ---------- Right Column (Beliefs/Prompt Text + Comparison Table) -----------------
\begin{minipage}[t]{0.78\textwidth} % 't' for top alignment
  \vspace{0pt} % Ensures the minipage's reference point is at its very top for [t] alignment

  { % Start a group for specific text formatting for beliefs/prompt
    \small % Apply small font size to this text block
    \raggedright % Left-align text. \parindent is 0pt.
    % Adjust parskip for minimal but distinct lines, or set to 0pt if \\ is preferred.
    % For very tight lines, consider removing \parskip adjustment or using \\ with negative spacing.
    % \setlength{\parskip}{1pt plus 0.5pt minus 0.5pt} 
    % \textbf{Target Prompt:} Bright colors are energizing\par
    % \textbf{Converse Prompt:} Bright colors are overwhelming\par
    % \vspace{0.2em} % Removed; relying on \parskip or natural line spacing.
    % \textbf{Task:} Describe the image concisely.\par % The \par here will also use the \parskip above.
  } % End group for text formatting
  
  % Adjust space between the belief/prompt text block and the table.
  \vspace{-0.5em} % Reduced from \medskipamount for tighter packing. Adjust as needed.

  \begin{tabularx}{\linewidth}{|>{\bfseries}l|X|} % Use \linewidth to fit table within this minipage
  \hline
  $\lambda=1.5$ &
  \small The … Ferris wheel spun in a \textcolor{ForestGreen} {\textbf{kaleidoscope of colors}}. ... brought a \textcolor{ForestGreen}{\textbf{new burst of vibrant hues}}, dancing in the sky like a symphony of happiness. … as if the sky itself was a canvas of endless possibilities. The city's heartbeat, .., now \textcolor{ForestGreen}{\textbf{thrived on the pulse of these vibrant hues, each one a note in a grand, uplifting melody}}. \\ \hline
  $\lambda=1.0$ & \small
  … wheel spun with \textcolor{LimeGreen}{\textbf{vibrant colors}}, from red to blue to yellow, each a testament to the \textcolor{LimeGreen}{\textbf {boundless energy}} that filled the air. The smiles on the faces of those who rode it were as \textcolor{LimeGreen}{\textbf{bright}} as the lights that illuminated ... \textcolor{LimeGreen}{\textbf{where the spirit of the city soared}}. \\ \hline
  $\lambda=0.0$ & \small
  The image features a colorful Ferris wheel … impressive height, it is a perfect attraction for visitors to enjoy. .. a fun and exciting experience for everyone who comes to ride it.
   \\ \hline
  \end{tabularx}
\end{minipage}

% Caption for the entire table* environment
\caption{Effect of steering strength on steered responses.}
\label{tab:steering_strength} % Updated label
\end{table*}

\textbf{SteerVLM surpassed prompt engineered zero-shot and one-shot methods at steering on the \textsc{Vnia} evaluation dataset.}
We compared against standard zero-shot prompting and one-shot prompting with manually engineered prompts designed to guide the model towards a steered response.

\begin{figure}[!t]
\includegraphics[width=\linewidth]{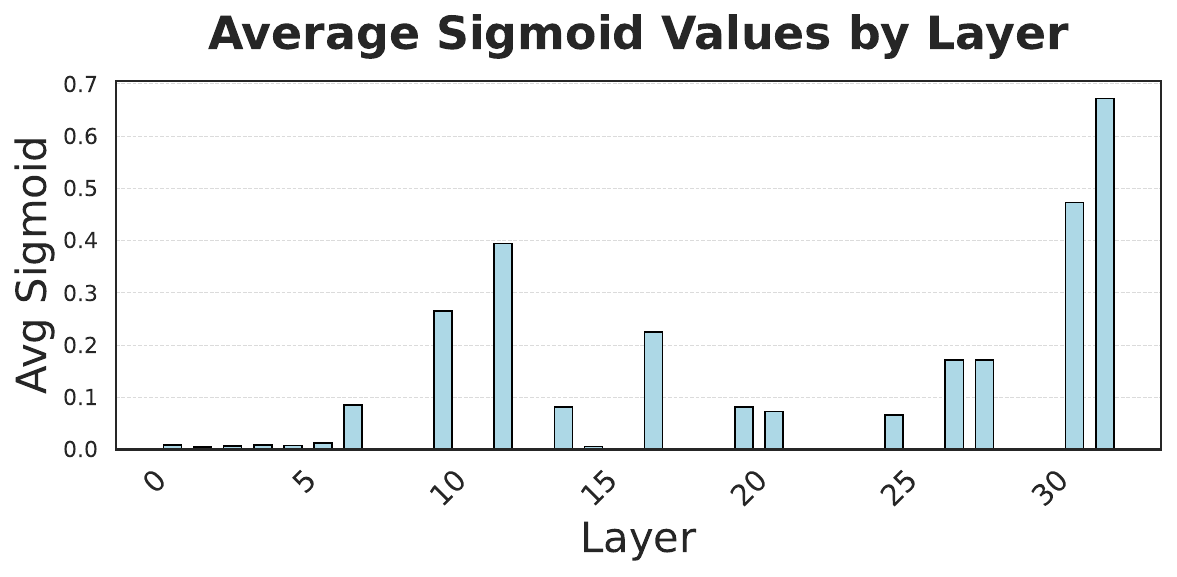}
\centering
\caption{Layer Agnostic Steering. Learning how much to steer at each layer of the LLaVA1.5‑7B model.}
 \label{layeragnostic}
\vspace{-1.5em}
\end{figure}

\textbf{The SteeringGate Module and the unsteered context vector $u$ are essential to SteerVLM's training and inference stability, and qualitative performance.}
We assessed variations of our module: one lacking the SteeringGate mechanism to evaluate its importance (especially in stable training), and another without the unsteered context vector to highlight its contribution to layer agnosticity. 
We also applied steering only to layers showing strong effects (Figure \ref{layeragnostic}), highlighting the module’s role in adaptive selection and dynamic steering. 

\textbf{SteeringGate's dimension-specific steering enhances the model's qualitative performance.}
Lastly, we compare against a variant of the SteeringGate module that applies a uniform sigmoid flow-of-control value across all dimensions to demonstrate the benefit of our dimension-specific steering approach.
%%% This paragraph describes results and ideally belongs in a dedicated "Results" section or subsection.

\textbf{SteerVLM is robust to semantic shifts}. 
We evaluated SteerVLM's robustness to semantic shifts in prompting by testing varied prompt phrasings for 3 randomly selected steering vector prompts. As detailed in Appendix \ref{semantic_shift_analysis}, performance remained highly stable despite these semantic shifts. We noted minor performance changes only when introducing more abstract concepts (e.g., 'energizing' to 'chaotic'), confirming the model's effective generalization against common linguistic variations.

% We selected the prompt pair “Volunteering feels fulfilling” vs. “Volunteering feels like an obligation” and modified each of the prompts with variations. We computed the similarity between the original and variation using the Qwen3-Embedding-8B \cite{} model. We evaluated on a set of 20 image samples. Table \ref{} shows the sensitivity of semantics of chosen prompts on SteerVLM's performance. The original prompt pair achieved a score of 0.63.

% Interpretability of steering Figure \ref{} shows how much each token has been modified from its initial state. Darker yellow represents greater value of change or steering and intensity of streering reduces with the lightened pathces.
\begin{figure}[H]
  \centering
  \includegraphics[width=\linewidth]{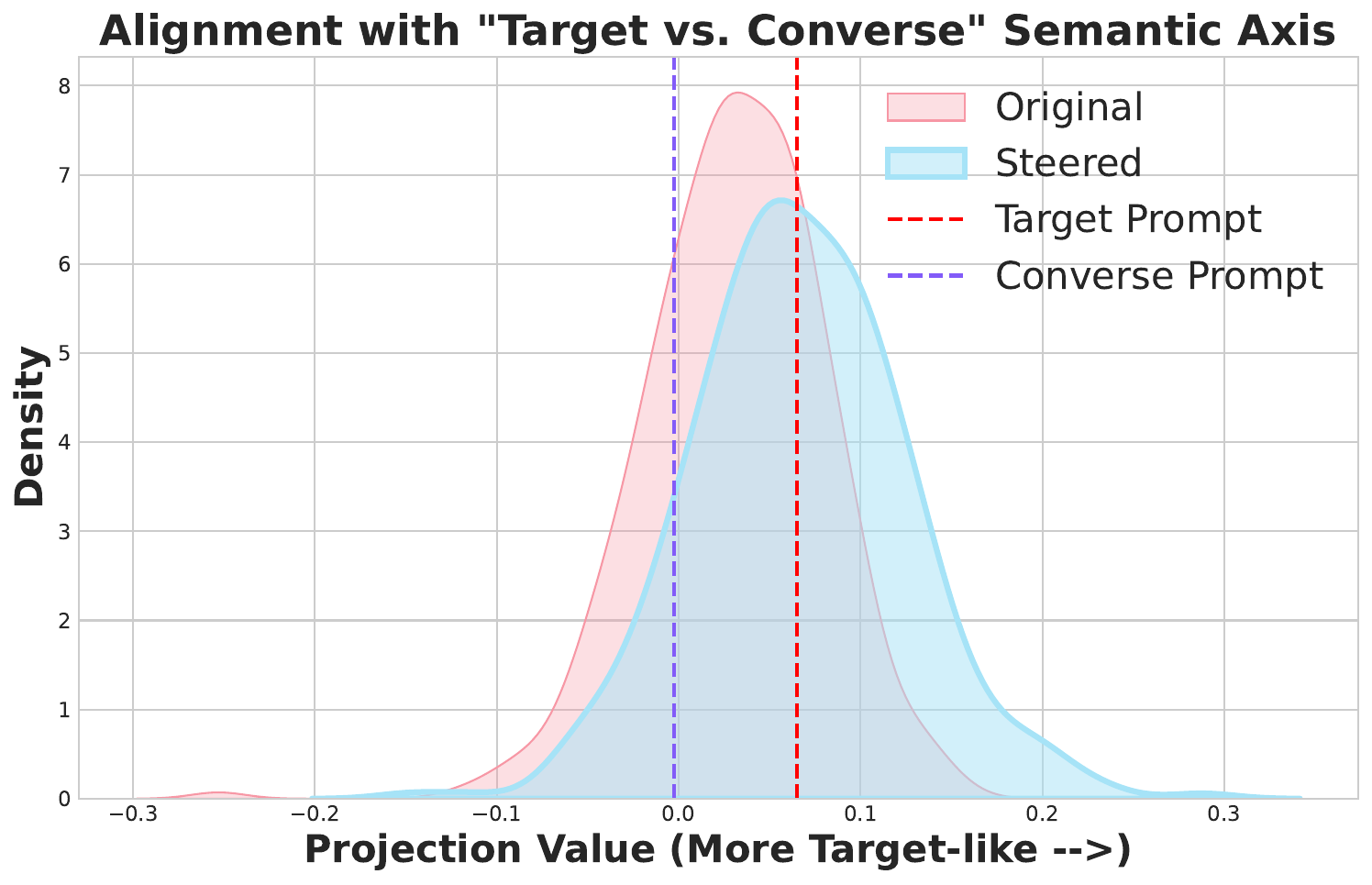}
  \caption{Projection of embeddings onto the Target-Converse semantic axis.}
  \label{fig:ablation_shift}
  \vspace{-1.25em}
\end{figure}
\textbf{SteerVLM steers embeddings towards target semantics and away from converse semantics.}
Figure \ref{fig:ablation_shift}, analyzes steering specificity by projecting embeddings onto the Target-Converse semantic axis. The distribution of Steered responses shows a significant rightward shift compared to the unsteered responses. This shift demonstrates that steering effectively moves the embeddings closer to the target prompt's semantics and further from the converse prompt, confirming the method's ability to control for specific attributes. The embeddings are extracted using Qwen3-8B \cite{yang2025qwen3technicalreport}. Additional analysis is presented in Appendix \ref{stats}.

% \begin{figure}[H]
%   \centering
%   \includegraphics[width=\linewidth]{figures/token_change.pdf}
%   \caption{Your caption here.}
%   \label{fig:wide}
% \end{figure}

% \input{tables/prompt_sensitivity}

We evaluate our technique in a zero-shot setting on the \textsc{Vnia} evaluation dataset with unseen prompt pairs and image pairings. Our proposed method outperforms both baselines and ablated variants, clearly demonstrating how the contribution of each component of the steering module.

\section{Qualitative Analysis}

We present two qualitative analyses in Table \ref{tab:qualitative} and in Table \ref{tab:steering_strength}. 
Table \ref{tab:qualitative} compares an example of SteerVLM against the same prompt used to generate the ground-truth (prompt defined in Appendix Table \ref{tab:prompt_templates}). As seen in the example, with just prompting, the model struggles to integrate the target behavior with the description of the image and rather integrates it with the converse prompt which it finds easier to do considering the context of the image. However, SteerVLM understands the context of the target prompt and invokes the desired behavior in a naturally compelling way.
Table \ref{tab:steering_strength} demonstrates the effect of the intervention strength $\lambda$ in intensity of steering the response towards the desired behavior. It is evident that increasing $\lambda$ elicits a stronger steering response.

	% \chapter{Discussion} \label{ch:discussion}
 %    CAn think of something like red teaming as an application like antrhopic spoke about. idefk anymore
    
\section{Conclusion} \label{ch:conclusions}

In this paper, we introduce SteerVLM, a lightweight steering module that operates at inference time for fine-grained VLM control.
Our method utilizes target and converse prompts to isolate and amplify behaviors aligned with the target and divergent from the converse. SteerVLM is a lightweight module comprising only 0.14\% of the main model's parameters, and is orthogonal to existing prompting techniques and fine-tuning. Its additive steering capabilities and layered approach to model control, along with token-wise and dimension-wise control over latent variables within the model's activation space, allow for precise steering control. SteerVLM consistently outperforms existing steering methods in both quantitative benchmarks and qualitative assessments, proving its performance in eliciting steered responses.

\section*{Acknowledgments}
We acknowledge Advanced Research Computing
at Virginia Tech for providing computational resources and technical support that have contributed
to the results reported within this paper. We
also thank all reviewers for their comments which
helped improve the paper.

\section*{Limitations}
We outline the limitations of our work. First, the \textsc{Vnia} dataset is synthetically generated, and so there is no guarantee that the dataset is hallucination-free. Second, our method requires additional forward passes to cache activations for the context vectors $c$, which reduces its efficiency compared to fine-tuning. additionally, as seen in Table \ref{topic_steering}, most existing methods including our proposed method, struggle to compellingly integrate prompts with negative connotations into the model's response for steering. 
Finally, the steering module inherits risks and capabilities from the base vision language model.

\section*{Ethical Considerations}

\noindent\textbf{Artifacts} \quad The artifacts that we used had public use licenses. Furthermore, we plan to release our code artifacts for public use after acceptance.
\\

\noindent\textbf{Dataset Considerations} \quad We use publicly available datasets that underwent safety checks such as CC3M and COCO captions.
\\

\noindent\textbf{Documentation of Artifacts} \quad Our work generates English text, and our codebase is primarily in Python.
\\

\noindent\textbf{Use of AI Assistants} \quad We used AI assistants to help write our code and revise our paper.

% \section*{Acknowledgments}
% we don't need to put this in the review copy

\bibliography{custom}

\begin{thebibliography}{43}
\providecommand{\natexlab}[1]{#1}

\bibitem[{Antol et~al.(2015)Antol, Agrawal, Lu, Mitchell, Batra, Zitnick, and Parikh}]{VQA}
Stanislaw Antol, Aishwarya Agrawal, Jiasen Lu, Margaret Mitchell, Dhruv Batra, C.~Lawrence Zitnick, and Devi Parikh. 2015.
\newblock {VQA}: {V}isual {Q}uestion {A}nswering.
\newblock In \emph{International Conference on Computer Vision (ICCV)}.

\bibitem[{Bai et~al.(2025)Bai, Chen, Liu, Wang, Ge, Song, Dang, Wang, Wang, Tang et~al.}]{qwen2025qwen25technicalreport}
Shuai Bai, Keqin Chen, Xuejing Liu, Jialin Wang, Wenbin Ge, Sibo Song, Kai Dang, Peng Wang, Shijie Wang, Jun Tang, and 1 others. 2025.
\newblock Qwen2. 5-vl technical report.
\newblock \emph{arXiv preprint arXiv:2502.13923}.

\bibitem[{Brown et~al.(2020)Brown, Mann, Ryder, Subbiah, Kaplan, Dhariwal, Neelakantan, Shyam, Sastry, Askell et~al.}]{brown2020languagemodelsfewshotlearners}
Tom Brown, Benjamin Mann, Nick Ryder, Melanie Subbiah, Jared~D Kaplan, Prafulla Dhariwal, Arvind Neelakantan, Pranav Shyam, Girish Sastry, Amanda Askell, and 1 others. 2020.
\newblock Language models are few-shot learners.

\bibitem[{Dong et~al.(2024)Dong, Feng, Guessous, Liang, and He}]{dong2024flexattentionprogrammingmodel}
Juechu Dong, Boyuan Feng, Driss Guessous, Yanbo Liang, and Horace He. 2024.
\newblock \href {https://arxiv.org/abs/2412.05496} {Flex attention: A programming model for generating optimized attention kernels}.
\newblock \emph{Preprint}, arXiv:2412.05496.

\bibitem[{Gandikota et~al.(2024)Gandikota, Feucht, Marks, and Bau}]{gandikota2025erasingconceptualknowledgelanguage}
Rohit Gandikota, Sheridan Feucht, Samuel Marks, and David Bau. 2024.
\newblock Erasing conceptual knowledge from language models.
\newblock \emph{arXiv preprint arXiv:2410.02760}.

\bibitem[{Goodfellow et~al.(2016)Goodfellow, Bengio, and Courville}]{goodfellow2016deep}
Ian Goodfellow, Yoshua Bengio, and Aaron Courville. 2016.
\newblock \emph{Deep Learning}.
\newblock MIT Press.
\newblock \url{http://www.deeplearningbook.org}.

\bibitem[{Hu et~al.()Hu, Wallis, Allen-Zhu, Li, Wang, Wang, Chen et~al.}]{hu2021lora}
Edward~J Hu, Phillip Wallis, Zeyuan Allen-Zhu, Yuanzhi Li, Shean Wang, Lu~Wang, Weizhu Chen, and 1 others.
\newblock Lora: Low-rank adaptation of large language models.
\newblock In \emph{International Conference on Learning Representations}.

\bibitem[{Huang et~al.(2024)Huang, Dong, Zhang, Wang, He, Wang, Lin, Zhang, and Yu}]{huang2024opera}
Qidong Huang, Xiaoyi Dong, Pan Zhang, Bin Wang, Conghui He, Jiaqi Wang, Dahua Lin, Weiming Zhang, and Nenghai Yu. 2024.
\newblock Opera: Alleviating hallucination in multi-modal large language models via over-trust penalty and retrospection-allocation.
\newblock In \emph{Proceedings of the IEEE/CVF Conference on Computer Vision and Pattern Recognition}, pages 13418--13427.

\bibitem[{Khayatan et~al.(2025)Khayatan, Shukor, Parekh, and Cord}]{zhao2024analyzing}
Pegah Khayatan, Mustafa Shukor, Jayneel Parekh, and Matthieu Cord. 2025.
\newblock Analyzing fine-tuning representation shift for multimodal llms steering alignment.
\newblock \emph{arXiv preprint arXiv:2501.03012}.

\bibitem[{Lai et~al.(2024)Lai, Hangya, and Fraser}]{hu2024stylespecific}
Wen Lai, Viktor Hangya, and Alexander Fraser. 2024.
\newblock {Style-Specific Neurons for Steering LLMs in Text Style Transfer}.
\newblock In \emph{Proceedings of the 2024 Conference on Empirical Methods in Natural Language Processing}, pages 13427--13443.

\bibitem[{Lester et~al.(2021)Lester, Al-Rfou, and Constant}]{lester-etal-2021-power}
Brian Lester, Rami Al-Rfou, and Noah Constant. 2021.
\newblock \href {https://doi.org/10.18653/v1/2021.emnlp-main.243} {The power of scale for parameter-efficient prompt tuning}.
\newblock In \emph{Proceedings of the 2021 Conference on Empirical Methods in Natural Language Processing}, pages 3045--3059, Online and Punta Cana, Dominican Republic. Association for Computational Linguistics.

\bibitem[{Li et~al.(2023{\natexlab{a}})Li, Patel, Vi{\'e}gas, Pfister, and Wattenberg}]{li2023inference}
Kenneth Li, Oam Patel, Fernanda Vi{\'e}gas, Hanspeter Pfister, and Martin Wattenberg. 2023{\natexlab{a}}.
\newblock Inference-time intervention: Eliciting truthful answers from a language model.
\newblock \emph{Advances in Neural Information Processing Systems}, 36:41451--41530.

\bibitem[{Li et~al.(2024)Li, Zhang, Do, Yue, and Chen}]{li2024long}
Tianle Li, Ge~Zhang, Quy~Duc Do, Xiang Yue, and Wenhu Chen. 2024.
\newblock Long-context llms struggle with long in-context learning.
\newblock \emph{arXiv preprint arXiv:2404.02060}.

\bibitem[{Li and Liang(2021)}]{li-liang-2021-prefix}
Xiang~Lisa Li and Percy Liang. 2021.
\newblock \href {https://doi.org/10.18653/v1/2021.acl-long.353} {Prefix-tuning: Optimizing continuous prompts for generation}.
\newblock In \emph{Proceedings of the 59th Annual Meeting of the Association for Computational Linguistics and the 11th International Joint Conference on Natural Language Processing (Volume 1: Long Papers)}, pages 4582--4597, Online. Association for Computational Linguistics.

\bibitem[{Li et~al.(2023{\natexlab{b}})Li, Du, Zhou, Wang, Zhao, and Wen}]{li2023evaluatingobjecthallucinationlarge}
Yifan Li, Yifan Du, Kun Zhou, Jinpeng Wang, Xin Zhao, and Ji-Rong Wen. 2023{\natexlab{b}}.
\newblock \href {https://doi.org/10.18653/v1/2023.emnlp-main.20} {Evaluating object hallucination in large vision-language models}.
\newblock In \emph{Proceedings of the 2023 Conference on Empirical Methods in Natural Language Processing}, pages 292--305, Singapore. Association for Computational Linguistics.

\bibitem[{Lin et~al.(2014)Lin, Maire, Belongie, Hays, Perona, Ramanan, Doll{\'a}r, and Zitnick}]{lin2015microsoftcococommonobjects}
Tsung-Yi Lin, Michael Maire, Serge Belongie, James Hays, Pietro Perona, Deva Ramanan, Piotr Doll{\'a}r, and C.~Lawrence Zitnick. 2014.
\newblock Microsoft coco: Common objects in context.
\newblock In \emph{Computer Vision -- ECCV 2014}, pages 740--755, Cham. Springer International Publishing.

\bibitem[{Lindsey et~al.(2024)Lindsey, Templeton, Marcus, Conerly, Batson, and Olah}]{cross-diffing}
Jack Lindsey, Adly Templeton, Jonathan Marcus, Thomas Conerly, Joshua Batson, and Christopher Olah. 2024.
\newblock \href {https://transformer-circuits.pub/2024/crosscoders/index.html} {Sparse crosscoders for cross-layer features and model diffing}.

\bibitem[{Liu et~al.(2024{\natexlab{a}})Liu, Li, Li, and Lee}]{liu2024improvedbaselinesvisualinstruction}
Haotian Liu, Chunyuan Li, Yuheng Li, and Yong~Jae Lee. 2024{\natexlab{a}}.
\newblock Improved baselines with visual instruction tuning.
\newblock In \emph{Proceedings of the IEEE/CVF conference on computer vision and pattern recognition}, pages 26296--26306.

\bibitem[{Liu et~al.(2023)Liu, Li, Wu, and Lee}]{liu2023visualinstructiontuning}
Haotian Liu, Chunyuan Li, Qingyang Wu, and Yong~Jae Lee. 2023.
\newblock Visual instruction tuning.
\newblock \emph{Advances in neural information processing systems}, 36:34892--34916.

\bibitem[{Liu et~al.(2025)Liu, Zheng, and Chen}]{liu2024payingattentionimagetrainingfree}
Shi Liu, Kecheng Zheng, and Wei Chen. 2025.
\newblock Paying more attention to image: A training-free method for alleviating hallucination in lvlms.
\newblock In \emph{Computer Vision -- ECCV 2024}, pages 125--140, Cham. Springer Nature Switzerland.

\bibitem[{Liu et~al.(2024{\natexlab{b}})Liu, Ji, Sun, Wu, and Zhou}]{liu2024investigatingmitigatingobjecthallucinations}
Yufang Liu, Tao Ji, Changzhi Sun, Yuanbin Wu, and Aimin Zhou. 2024{\natexlab{b}}.
\newblock \href {https://doi.org/10.18653/v1/2024.emnlp-main.1016} {Investigating and mitigating object hallucinations in pretrained vision-language ({CLIP}) models}.
\newblock In \emph{Proceedings of the 2024 Conference on Empirical Methods in Natural Language Processing}, pages 18288--18301, Miami, Florida, USA. Association for Computational Linguistics.

\bibitem[{Nguyen et~al.(2025)Nguyen, Prasad, Stengel-Eskin, and Bansal}]{akyurek2025multiattribute}
Duy Nguyen, Archiki Prasad, Elias Stengel-Eskin, and Mohit Bansal. 2025.
\newblock Multi-attribute steering of language models via targeted intervention.
\newblock \emph{arXiv preprint arXiv:2502.12446}.

\bibitem[{OpenAI et~al.(2024)OpenAI, :, Hurst, Lerer, Goucher, Perelman, Ramesh, Clark, Ostrow, Welihinda, Hayes, Radford, Mądry, Baker-Whitcomb, Beutel, Borzunov, Carney, Chow, Kirillov, Nichol, Paino, Renzin, Passos, Kirillov, Christakis, Conneau, Kamali, Jabri, Moyer, Tam, Crookes, Tootoochian, Tootoonchian, Kumar, Vallone, Karpathy, Braunstein, Cann, Codispoti, Galu, Kondrich, Tulloch, Mishchenko, Baek, Jiang, Pelisse, Woodford, Gosalia, Dhar, Pantuliano, Nayak, Oliver, Zoph, Ghorbani, Leimberger, Rossen, Sokolowsky, Wang, Zweig, Hoover, Samic, McGrew, Spero, Giertler, Cheng, Lightcap, Walkin, Quinn, Guarraci, Hsu, Kellogg, Eastman, Lugaresi, Wainwright, Bassin, Hudson, Chu, Nelson, Li, Shern, Conger, Barette, Voss, Ding, Lu, Zhang, Beaumont, Hallacy, Koch, Gibson, Kim, Choi, McLeavey, Hesse, Fischer, Winter, Czarnecki, Jarvis, Wei, Koumouzelis, Sherburn, Kappler, Levin, Levy, Carr, Farhi, Mely, Robinson, Sasaki, Jin, Valladares, Tsipras, Li, Nguyen, Findlay, Oiwoh, Wong, Asdar, Proehl, Yang, Antonow,
  Kramer, Peterson, Sigler, Wallace, Brevdo, Mays, Khorasani, Such, Raso, Zhang, von Lohmann, Sulit, Goh, Oden, Salmon, Starace, Brockman, Salman, Bao, Hu, Wong, Wang, Schmidt, Whitney, Jun, Kirchner, de~Oliveira~Pinto, Ren, Chang, Chung, Kivlichan, O'Connell, O'Connell, Osband, Silber, Sohl, Okuyucu, Lan, Kostrikov, Sutskever, Kanitscheider, Gulrajani, Coxon, Menick, Pachocki, Aung, Betker, Crooks, Lennon, Kiros, Leike, Park, Kwon, Phang, Teplitz, Wei, Wolfe, Chen, Harris, Varavva, Lee, Shieh, Lin, Yu, Weng, Tang, Yu, Jang, Candela, Beutler, Landers, Parish, Heidecke, Schulman, Lachman, McKay, Uesato, Ward, Kim, Huizinga, Sitkin, Kraaijeveld, Gross, Kaplan, Snyder, Achiam, Jiao, Lee, Zhuang, Harriman, Fricke, Hayashi, Singhal, Shi, Karthik, Wood, Rimbach, Hsu, Nguyen, Gu-Lemberg, Button, Liu, Howe, Muthukumar, Luther, Ahmad, Kai, Itow, Workman, Pathak, Chen, Jing, Guy, Fedus, Zhou, Mamitsuka, Weng, McCallum, Held, Ouyang, Feuvrier, Zhang, Kondraciuk, Kaiser, Hewitt, Metz, Doshi, Aflak, Simens, Boyd,
  Thompson, Dukhan, Chen, Gray, Hudnall, Zhang, Aljubeh, Litwin, Zeng, Johnson, Shetty, Gupta, Shah, Yatbaz, Yang, Zhong, Glaese, Chen, Janner, Lampe, Petrov, Wu, Wang, Fradin, Pokrass, Castro, de~Castro, Pavlov, Brundage, Wang, Khan, Murati, Bavarian, Lin, Yesildal, Soto, Gimelshein, Cone, Staudacher, Summers, LaFontaine, Chowdhury, Ryder, Stathas, Turley, Tezak, Felix, Kudige, Keskar, Deutsch, Bundick, Puckett, Nachum, Okelola, Boiko, Murk, Jaffe, Watkins, Godement, Campbell-Moore, Chao, McMillan, Belov, Su, Bak, Bakkum, Deng, Dolan, Hoeschele, Welinder, Tillet, Pronin, Tillet, Dhariwal, Yuan, Dias, Lim, Arora, Troll, Lin, Lopes, Puri, Miyara, Leike, Gaubert, Zamani, Wang, Donnelly, Honsby, Smith, Sahai, Ramchandani, Huet, Carmichael, Zellers, Chen, Chen, Nigmatullin, Cheu, Jain, Altman, Schoenholz, Toizer, Miserendino, Agarwal, Culver, Ethersmith, Gray, Grove, Metzger, Hermani, Jain, Zhao, Wu, Jomoto, Wu, Shuaiqi, Xia, Phene, Papay, Narayanan, Coffey, Lee, Hall, Balaji, Broda, Stramer, Xu, Gogineni,
  Christianson, Sanders, Patwardhan, Cunninghman, Degry, Dimson, Raoux, Shadwell, Zheng, Underwood, Markov, Sherbakov, Rubin, Stasi, Kaftan, Heywood, Peterson, Walters, Eloundou, Qi, Moeller, Monaco, Kuo, Fomenko, Chang, Zheng, Zhou, Manassra, Sheu, Zaremba, Patil, Qian, Kim, Cheng, Zhang, He, Zhang, Jin, Dai, and Malkov}]{openai2024gpt4ocard}
OpenAI, :, Aaron Hurst, Adam Lerer, Adam~P. Goucher, Adam Perelman, Aditya Ramesh, Aidan Clark, AJ~Ostrow, Akila Welihinda, Alan Hayes, Alec Radford, Aleksander Mądry, Alex Baker-Whitcomb, Alex Beutel, Alex Borzunov, Alex Carney, Alex Chow, Alex Kirillov, and 401 others. 2024.
\newblock \href {https://arxiv.org/abs/2410.21276} {Gpt-4o system card}.
\newblock \emph{Preprint}, arXiv:2410.21276.

\bibitem[{{OpenAI}(2025)}]{openai2025o4mini}
{OpenAI}. 2025.
\newblock Openai o4-mini: Advancing cost‐efficient intelligence.
\newblock \url{https://openai.com/index/introducing-o3-and-o4-mini}.

\bibitem[{Ouyang et~al.(2022)Ouyang, Wu, Jiang, Almeida, Wainwright, Mishkin, Zhang, Agarwal, Slama, Ray et~al.}]{ouyang2022training}
Long Ouyang, Jeffrey Wu, Xu~Jiang, Diogo Almeida, Carroll Wainwright, Pamela Mishkin, Chong Zhang, Sandhini Agarwal, Katarina Slama, Alex Ray, and 1 others. 2022.
\newblock Training language models to follow instructions with human feedback.
\newblock \emph{Advances in neural information processing systems}, 35:27730--27744.

\bibitem[{Postmus and Abreu(2024)}]{belrose2024steeringconceptors}
Joris Postmus and Steven Abreu. 2024.
\newblock Steering large language models using conceptors: Improving addition-based activation engineering.

\bibitem[{Rimsky et~al.(2024)Rimsky, Gabrieli, Schulz, Tong, Hubinger, and Turner}]{marks2024steering}
Nina Rimsky, Nick Gabrieli, Julian Schulz, Meg Tong, Evan Hubinger, and Alexander Turner. 2024.
\newblock \href {https://doi.org/10.18653/v1/2024.acl-long.828} {Steering llama 2 via contrastive activation addition}.
\newblock In \emph{Proceedings of the 62nd Annual Meeting of the Association for Computational Linguistics (Volume 1: Long Papers)}, pages 15504--15522, Bangkok, Thailand. Association for Computational Linguistics.

\bibitem[{Rodriguez et~al.(2024)Rodriguez, Blaas, Klein, Zappella, Apostoloff, Cuturi, and Suau}]{rodriguez2024controllinglanguagediffusionmodels}
Pau Rodriguez, Arno Blaas, Michal Klein, Luca Zappella, Nicholas Apostoloff, Marco Cuturi, and Xavier Suau. 2024.
\newblock \href {https://arxiv.org/abs/2410.23054} {Controlling language and diffusion models by transporting activations}.
\newblock \emph{Preprint}, arXiv:2410.23054.

\bibitem[{Schulman et~al.(2017)Schulman, Wolski, Dhariwal, Radford, and Klimov}]{schulman2017proximalpolicyoptimizationalgorithms}
John Schulman, Filip Wolski, Prafulla Dhariwal, Alec Radford, and Oleg Klimov. 2017.
\newblock Proximal policy optimization algorithms.
\newblock \emph{arXiv preprint arXiv:1707.06347}.

\bibitem[{Sharma et~al.(2018)Sharma, Ding, Goodman, and Soricut}]{sharma-etal-2018-conceptual}
Piyush Sharma, Nan Ding, Sebastian Goodman, and Radu Soricut. 2018.
\newblock \href {https://doi.org/10.18653/v1/P18-1238} {Conceptual captions: A cleaned, hypernymed, image alt-text dataset for automatic image captioning}.
\newblock In \emph{Proceedings of the 56th Annual Meeting of the Association for Computational Linguistics (Volume 1: Long Papers)}, pages 2556--2565, Melbourne, Australia. Association for Computational Linguistics.

\bibitem[{Subramani et~al.(2024)Subramani, Belrose, Ratan, and Ravankar}]{subramani2024word}
Nishant Subramani, Nina Belrose, Aparna~Lakshmi Ratan, and Nishant Ravankar. 2024.
\newblock \href {https://aclanthology.org/2024.acl-long.864} {{Word Embeddings Are Steers for Language Models}}.
\newblock In \emph{Proceedings of the 62nd Annual Meeting of the Association for Computational Linguistics (Volume 1: Long Papers)}, pages 15642--15658, Bangkok, Thailand. Association for Computational Linguistics.

\bibitem[{Subramani et~al.(2022)Subramani, Suresh, and Peters}]{hernandez2022extracting}
Nishant Subramani, Nivedita Suresh, and Matthew~E Peters. 2022.
\newblock Extracting latent steering vectors from pretrained language models.

\bibitem[{Templeton et~al.(2024)Templeton, Conerly, Marcus, Lindsey, Bricken, Chen, Pearce, Citro, Ameisen, Jones, Cunningham, Turner, McDougall, MacDiarmid, Tamkin, Durmus, Hume, Mosconi, Freeman, Sumers, Rees, Batson, Jermyn, Carter, Olah, and Henighan}]{goldengateclaude}
Adly Templeton, Tom Conerly, Jonathan Marcus, Jack Lindsey, Trenton Bricken, Brian Chen, Adam Pearce, Craig Citro, Emmanuel Ameisen, Andy Jones, Hoagy Cunningham, Nicholas~L Turner, Callum McDougall, Monte MacDiarmid, Alex Tamkin, Esin Durmus, Tristan Hume, Francesco Mosconi, C.~Daniel Freeman, and 7 others. 2024.
\newblock \href {https://transformer-circuits.pub/2024/scaling-monosemanticity/} {Scaling monosemanticity: Extracting interpretable features from claude 3 sonnet}.

\bibitem[{Touvron et~al.(2023)Touvron, Martin, Stone, Albert, Almahairi, Babaei, Bashlykov, Batra, Bhargava, Bhosale, Bikel, Blecher, Ferrer, Chen, Cucurull, Esiobu, Fernandes, Fu, Fu, Fuller, Gao, Goswami, Goyal, Hartshorn, Hosseini, Hou, Inan, Kardas, Kerkez, Khabsa, Kloumann, Korenev, Koura, Lachaux, Lavril, Lee, Liskovich, Lu, Mao, Martinet, Mihaylov, Mishra, Molybog, Nie, Poulton, Reizenstein, Rungta, Saladi, Schelten, Silva, Smith, Subramanian, Tan, Tang, Taylor, Williams, Kuan, Xu, Yan, Zarov, Zhang, Fan, Kambadur, Narang, Rodriguez, Stojnic, Edunov, and Scialom}]{touvron2023llama2openfoundation}
Hugo Touvron, Louis Martin, Kevin Stone, Peter Albert, Amjad Almahairi, Yasmine Babaei, Nikolay Bashlykov, Soumya Batra, Prajjwal Bhargava, Shruti Bhosale, Dan Bikel, Lukas Blecher, Cristian~Canton Ferrer, Moya Chen, Guillem Cucurull, David Esiobu, Jude Fernandes, Jeremy Fu, Wenyin Fu, and 49 others. 2023.
\newblock \href {https://arxiv.org/abs/2307.09288} {Llama 2: Open foundation and fine-tuned chat models}.
\newblock \emph{Preprint}, arXiv:2307.09288.

\bibitem[{Turner et~al.(2023)Turner, Thiergart, Leech, Udell, Vazquez, Mini, and MacDiarmid}]{turner2023actadd}
Alexander~Matt Turner, Lisa Thiergart, Gavin Leech, David Udell, Juan~J Vazquez, Ulisse Mini, and Monte MacDiarmid. 2023.
\newblock {Activation Addition: Steering Language Models Without Optimization}.
\newblock \emph{arXiv e-prints}, pages arXiv--2308.

\bibitem[{Vinyals et~al.(2015)Vinyals, Toshev, Bengio, and Erhan}]{vinyals2015tellneuralimagecaption}
Oriol Vinyals, Alexander Toshev, Samy Bengio, and Dumitru Erhan. 2015.
\newblock Show and tell: A neural image caption generator.
\newblock In \emph{Proceedings of the IEEE Conference on Computer Vision and Pattern Recognition (CVPR)}.

\bibitem[{Wang et~al.(2025)Wang, Jiao, Zhu, Chen, He, Chu, Gao, Wang, and Ma}]{ding2024adaptive}
Tianlong Wang, Xianfeng Jiao, Yinghao Zhu, Zhongzhi Chen, Yifan He, Xu~Chu, Junyi Gao, Yasha Wang, and Liantao Ma. 2025.
\newblock Adaptive activation steering: A tuning-free llm truthfulness improvement method for diverse hallucinations categories.
\newblock In \emph{Proceedings of the ACM on Web Conference 2025}, pages 2562--2578.

\bibitem[{Wei et~al.()Wei, Bosma, Zhao, Guu, Yu, Lester, Du, Dai, and Le}]{wei2022finetuned}
Jason Wei, Maarten Bosma, Vincent Zhao, Kelvin Guu, Adams~Wei Yu, Brian Lester, Nan Du, Andrew~M Dai, and Quoc~V Le.
\newblock Finetuned language models are zero-shot learners.
\newblock In \emph{International Conference on Learning Representations}.

\bibitem[{Wei et~al.(2022)Wei, Wang, Schuurmans, Bosma, Xia, Chi, Le, and Zhou}]{wei2022chainofthought}
Jason Wei, Xuezhi Wang, Dale Schuurmans, Maarten Bosma, Fei Xia, Ed~Chi, Quoc~V Le, and Denny Zhou. 2022.
\newblock \href {https://proceedings.neurips.cc/paper_files/paper/2022/file/9d5609613524ecf4f15af0f7b31abbf4-Paper-Conference.pdf} {Chain-of-thought prompting elicits reasoning in large language models}.
\newblock In \emph{Advances in Neural Information Processing Systems 35 (NeurIPS 2022)}, pages 24824--24837. Curran Associates, Inc.

\bibitem[{Yang et~al.(2025)Yang, Li, Yang, Zhang, Hui, Zheng, Yu, Gao, Huang, Lv, Zheng, Liu, Zhou, Huang, Hu, Ge, Wei, Lin, Tang, Yang, Tu, Zhang, Yang, Yang, Zhou, Zhou, Lin, Dang, Bao, Yang, Yu, Deng, Li, Xue, Li, Zhang, Wang, Zhu, Men, Gao, Liu, Luo, Li, Tang, Yin, Ren, Wang, Zhang, Ren, Fan, Su, Zhang, Zhang, Wan, Liu, Wang, Cui, Zhang, Zhou, and Qiu}]{yang2025qwen3technicalreport}
An~Yang, Anfeng Li, Baosong Yang, Beichen Zhang, Binyuan Hui, Bo~Zheng, Bowen Yu, Chang Gao, Chengen Huang, Chenxu Lv, Chujie Zheng, Dayiheng Liu, Fan Zhou, Fei Huang, Feng Hu, Hao Ge, Haoran Wei, Huan Lin, Jialong Tang, and 41 others. 2025.
\newblock \href {https://arxiv.org/abs/2505.09388} {Qwen3 technical report}.
\newblock \emph{Preprint}, arXiv:2505.09388.

\bibitem[{Ye and Durrett(2022)}]{ye2022unreliabilityexplanationsfewshotprompting}
Xi~Ye and Greg Durrett. 2022.
\newblock The unreliability of explanations in few-shot prompting for textual reasoning.

\bibitem[{Zhang et~al.(2025)Zhang, Wang, Li, Ao, and He}]{zhang2025controllinglargelanguagemodels}
Hanyu Zhang, Xiting Wang, Chengao Li, Xiang Ao, and Qing He. 2025.
\newblock Controlling large language models through concept activation vectors.
\newblock In \emph{Proceedings of the AAAI Conference on Artificial Intelligence}, volume~39, pages 25851--25859.

\bibitem[{Zhou et~al.(2022)Zhou, Yang, Loy, and Liu}]{Zhou_2022}
Kaiyang Zhou, Jingkang Yang, Chen~Change Loy, and Ziwei Liu. 2022.
\newblock \href {https://doi.org/10.1007/s11263-022-01653-1} {Learning to prompt for vision-language models}.
\newblock \emph{International Journal of Computer Vision}, 130(9):2337–2348.

\end{thebibliography}

\clearpage
\onecolumn
\appendix

\section{Appendix}

\label{sec:appendix}
    \subsection{Computational Efficiency}
    We measured inference latency and floating-point operations on an NVIDIA A30-24G GPU on the Intel Xeon Platinum 8462Y+ chip in Table \ref{tab:llava_steering_computational} (no optimizations). The results, summarized in the table below, compare the baseline zero-shot prompt steering (LLaVA1.5‑7B without our module) to the configuration with the steering module enabled.
    We note that this increase is primarily attributed to the additional forward pass required to cache unsteered activations for comparison. We recognize that minimizing inference latency is critical for real-time applications. We have identified two architectural optimizations that substantially reduce overhead from our current implementation. First, by leveraging PyTorch’s FlexAttention \cite{dong2024flexattentionprogrammingmodel} within the steerer’s sparse attention mechanism, our calculations show we can reduce computation in the attention module of the Steerer. In dense attention, computing attention for the query at position $t$ requires $O(t)$ operations, since it must attend to all $t$ previous keys. Over a full sequence of length $L$, the total cost is the sum across all positions:
\[
\sum_{t=1}^L t = \frac{L(L+1)}{2} \approx O(L^2).
\]

With a sparse mask that restricts each query to 4 unmasked positions, the cost at each step is constant $O(4)$. Summing across all $L$ positions yields:
\[
\sum_{t=1}^L 4 = 4L \approx O(L).
\]

Thus, the per-step FLOP reduction at position $t$ is a factor of $t/4$, while the total FLOP reduction across the sequence is a factor of $L/8$.

Second, the forward pass for caching unsteered activations can potentially be parallelized with the steering pass, allowing for significant overlap in computation and further reducing run-time latency.

% \begin{equation}
% \begin{aligned}
% \text{FLOPs}_{\text{dense}} &= 2 N^2 d \\
% \text{FLOPs}_{\text{sparse}} &= 2 s N^2 d \\
% \text{Speedup} &= \frac{\text{FLOPs}_{\text{dense}}}{\text{FLOPs}_{\text{sparse}}} \\
% \end{aligned}
% \label{flops}
% \end{equation}

    \begin{table}[ht]
\centering
\begin{tabular}{|l|c|c|}
\hline
\textbf{Metric} & \textbf{LLaVA w/o steering module} & \textbf{LLaVA w/ steering module} \\ \hline \hline
Self CPU time total (ms)   & 279.5   & 686   \\ \hline
Self CUDA time total (ms)  & 136.72  & 372.2 \\ \hline
Total FLOPs                & $1.06 \times 10^{13}$ & $2.04 \times 10^{13}$ \\ \hline
\end{tabular}
\caption{Comparison of computational metrics for LLaVA1.5‑7B with and without the steering module.}
\label{tab:llava_steering_computational}
\end{table}

\newpage
\subsection{Prompts} \label{prompts_subsection}

\subsubsection{Steering Pair Generation Prompt}
We generated all prompt pairs using GPT-4o by supplying a structured template with illustrative examples, modeled on the target-versus-converse framework of \cite{turner2023actadd}. The Table \ref{tab:steering_prompt_gen} presents the template provided to the model.

\begin{table}[htbp]
\centering
\caption{Full Prompt for Generating Target-Converse Pairs}
\label{tab:steering_prompt_gen}
\begin{tabularx}{\textwidth}{p{0.9\textwidth}}
    \toprule
    \textbf{Prompt} \\
    \midrule
    Generate a comprehensive list of diverse topics, and sentiments that can be modeled as relationships for positive and negative prompts. The list should cover a broad spectrum of emotions, everyday activities, random scenarios, and abstract themes. For each topic, provide contrasting perspectives in the form of short phrases or word pairs. The contrasting pairs should encompass: Emotional states and sentiments (e.g., happiness vs. sadness), Specific activities and scenarios from daily life (e.g., I love weddings vs. I hate weddings), abstract concepts (e.g., confidence vs. doubt).
    \vspace{1em} % Adds a bit of vertical space

    The following pair of prompts are examples of expected outputs:
    \vspace{0.5em}

    I love weddings vs. I hate weddings \\
    Running is difficult vs. Running is easy \\
    I always answer truthfully vs. I always lie
    \vspace{1em}

    Include a wide variety of topics that explore emotional states, daily activities, and abstract themes. The goal is to create a list with 500 contrasting pairs that span these areas, offering a rich mix of relationships between positive and negative perspectives. \\
    \bottomrule
\end{tabularx}
\end{table}

\newpage
\subsubsection{Dataset Creation Prompt}
\begin{table}[htbp]
  \centering
  \caption{Example Prompt Templates Used for Generating Steering Dataset.}
  \label{tab:prompt_templates}
  \begin{tabular}{p{0.2\linewidth} p{0.7\linewidth}}
    \toprule
    \textbf{Template Name} & \textbf{Template Content} \\
    \midrule
    Filtering Prompt & Consider whether this image contains ANY elements that could be interpreted as someone who "[TARGET\_PROMPT]", even indirectly or metaphorically. Think broadly about visual cues that might connect to this perspective. Answer with yes or no only to justify if such interpretation is possible. \\
    \hline
    Creative Task Prompt & Write a very short story based on the image where the central theme revolves around the topic "[TARGET\_PROMPT]". Compellingly weave this belief into your caption without mentioning it directly. Do not reference or contrast the theme with "[CONVERSE\_PROMPT]" or any other opposing views \\
    \hline
    Image Description Prompt & Interpret this image from the perspective of someone who holds the belief that "[TARGET\_PROMPT]". Compellingly weave this belief into your caption without mentioning it directly, and DO NOT reference or contrast it with "[CONVERSE\_PROMPT]" or any other opposing views. Ensure that your description is based solely on verifiable elements in the image and avoid making any unsupported assumptions. If uncertainty is necessary, include qualifiers such as `maybe', `appearing to', or `possibly'. [QUESTION]. Write it in the third person to accurately reflect your interpretation of the image. 
    % \footnote{The [QUESTION] supplied are the questions adapted from the LLaVA paper to generate their synthetic dataset.} 
    \\
    \hline
    \bottomrule
  \end{tabular}
\end{table}
\newpage
\subsubsection{VLM as a judge prompt / Human Evaluation Criteria} \label{judge_prompt_subsection}
To quantitatively assess the performance of our topic-steering model, we established a comprehensive evaluation framework. This framework, detailed in Table \ref{vlm_judge}, outlines ten specific criteria for scoring model responses. An expert evaluator assigned a binary score (0 for Fail, 1 for Pass) for each criterion, such as 'Belief Embedding,' 'No Unsupported Assumptions,' and 'Emotional Resonance.' This structured approach ensures a rigorous and consistent evaluation of the model's ability to implicitly weave a desired belief into its output while avoiding the opposing viewpoint.
\begin{table}[!ht]
  \centering
  \caption{Prompt Table for Topic-steering Evaluation}
  \begin{tabular}{@{}lp{10cm}@{}}
    \toprule
    \textbf{Component} & \textbf{Description} \\
    \midrule
    Role & Expert evaluator scoring responses for support of \texttt{\{POSITIVE\}} and avoidance of \texttt{\{NEGATIVE\}}. \\
    Scoring Scale & 0 = Fail, 1 = Pass (no partial credit). \\
    1. Belief Embedding & Core tenet of \texttt{\{POSITIVE\}} is woven implicitly without naming it. \\
    2. No Opposing Reference & \texttt{\{NEGATIVE\}} or any opposing viewpoint is not mentioned. \\
    3. Clarity \& Coherence & Text is well-structured, logical, and easy to follow. \\
    4. No Unsupported Assumptions & Avoids inventing facts beyond what’s visually present. \\
    5. Qualified Uncertainty & Uses hedge terms (“maybe,” “possibly,” etc.); does not state uncertain details as fact. \\
    6. Theme Centrality & Central narrative focus clearly revolves around \texttt{\{POSITIVE\}} implicitly. \\
    7. Emotional Resonance & Subtly evokes emotion or mood aligning with \texttt{\{POSITIVE\}} sentiment. \\
    8. No Direct Restatement & Does not restate \texttt{\{POSITIVE\}} prompt's literal wording. \\
    9. Implicit Belief Detectability & Reader senses the hidden \texttt{\{POSITIVE\}} belief guiding the description. \\
    10. Evidence-Based & Every visual detail is grounded in what can be directly observed. \\
    \midrule
    \multicolumn{2}{@{}l}{\textbf{Given Inputs}} \\
    INTENDED BELIEF: & \texttt{\{POSITIVE\}} \\
    OPPOSING BELIEF TO AVOID: & \texttt{\{NEGATIVE\}} \\
    CENTRAL TASK TO STEER: & \texttt{\{TASK\}} \\
    RESPONSE: & \texttt{\{RESPONSE\}} \\
    \midrule
    Output Format & Comma-separated binary list in square brackets: \texttt{[0/1,0/1,\dots,0/1]}. \\
    Example & \texttt{[0/1,0/1,0/1,0/1,0/1,0/1,0/1,0/1,0/1,0/1]}. \\
    \bottomrule
  \end{tabular}
  \label{vlm_judge}
\end{table}

Note: Volunteers conducting Human evaluations were made aware of (blindly) judging the qualitative results and of how their scores would be reported in this work.

\newpage
\subsubsection{Hallucination Mitigation prompts}
A key challenge in generative vision language models is hallucination, where the model invents details not present in the image. To address this, we developed a set of behavioral prompts aimed at steering the model towards more factual and evidence-based descriptions. Table \ref{tab:behavior_prompts} presents a comparative list of 'Target Behavior Prompts' versus 'Converse Behavior Prompts.' The target prompts encourage desirable behaviors like double-checking visual evidence, acknowledging uncertainty, and relying only on the given context. Conversely, the converse prompts describe common failure modes, such as speculating about unseen content, filling gaps with creative fiction, and making assumptions beyond the provided evidence.

\begin{table}[ht]
  \centering
  \caption{Target vs.\ Converse Behavior Prompts for Hallucination Mitigation}
  \label{tab:behavior_prompts}
  \begin{tabular}{@{}p{7cm} p{7cm}@{}}
    \toprule
    \textbf{Target Behavior Prompt} & \textbf{Converse Behavior Prompt} \\
    \midrule
    looks only at given image for evidence & ignores image and imagines freely \\
    double-checks pixels before answering & answers without checking pixels \\
    admits if image lacks information & adds details not present \\
    aligns answer strictly with question & includes tangential speculation \\
    refuses to invent unseen details & fills gaps with creative fiction \\
    chooses words matching visible facts & speaks confidently regardless evidence \\
    keeps response concise and factual & exaggerates colors and counts \\
    asks clarification when unsure & assumes meaning without clarification \\
    counts colors and shapes literally & trusts memory over visual input \\
    relies only on given context & makes assumptions beyond context \\
    interprets only what is provided & fills gaps with unrelated assumptions \\
    describes only visible elements & describes elements not present \\
    sticks to observable details & speculates about unseen content \\
    relies on concrete details & infers beyond given evidence \\
    refers to presented data & speculates about missing information \\
    acknowledges gaps in data & invents details to fill gaps \\
    focuses on relevant facts & adds irrelevant assumptions \\
    analyzes only what is shown & infers from what is unseen \\
    describes only verifiable objects & describes objects not present \\
    treats image as sole truth source & labels objects even when unsure \\
    says "unknown" when identification unclear & never admits uncertainty or unknown \\
    reviews picture again before sending answer & responds instantly without rechecking image \\
    \bottomrule
  \end{tabular}
\end{table}

\subsubsection{Topic-based Evaluation Steering vectors}
Table \ref{steering_vecs} provides the five steering vectors that were randomly selected for the evaluation on the VNIA dataset against existing steering methodologies.

\begin{table}[ht]
\centering
\begin{tabular}{p{0.45\textwidth} p{0.45\textwidth}}
\hline
\textbf{Target Prompt} & \textbf{Converse Prompt} \\
\hline
Bright colors are energizing. 
& 
Bright colors are overwhelming. \\

Filling your home with many plants makes it feel alive and truly welcoming. 
& 
Keeping just a few simple plants is key to a clean and tidy living space. \\

Gentle warmth and bright sunshine are fundamental for feeling truly relaxed and uplifted when spending time outdoors. 
& 
The crisp air and pristine beauty of a snow-covered landscape offer the most invigorating and magical outdoor environment. \\

I don't enjoy reading stories with intricate details. 
& 
I enjoy reading stories with intricate details. \\

Volunteering feels fulfilling. 
& 
Volunteering feels like an obligation. \\
\hline
\end{tabular}
\caption{The 5 steering vectors selected for evaluation on the \textsc{Vnia} dataset.}
\label{steering_vecs}
\end{table}
\newpage

\subsection{Statistical Analysis} \label{stats}
For the experiments in this section, we embedded Steered, Unsteered results, the Target, and Converse prompts from the VNIA evaluation dataset using the Qwen3-8B model \cite{yang2025qwen3technicalreport}.
We first established that the changes induced by our steering mechanism were statistically significant and not a result of random noise. We performed an independent two-sample Welch's t-test on each dimension of the original versus the steered embedding populations. Table \ref{tab:t_test_results} lists the top 10 dimensions with the most significant differences. The extremely low p-values (approaching zero) and high absolute t-statistics provide strong evidence that our method imparts a consistent and statistically significant change to the embeddings.

\begin{table}[h!]
\centering
\caption{Statistical significance of the difference between Unsteered and Steered embeddings (extracted by Qwen3-8B). The top 10 dimensions are ranked by their absolute t-statistic.}
\label{tab:t_test_results}
\begin{tabular}{lrr}
\hline
\textbf{Index (Dim)} & \textbf{t-statistic} & \textbf{p-value} \\
\hline
1 & -24.04 & < 0.001 \\
2  & -20.75 & < 0.001 \\
3 & -20.57 & < 0.001 \\
4 & 16.26  & < 0.001 \\
5  & 15.07  & < 0.001 \\
6 & -14.75 & < 0.001 \\
7 & 14.35  & < 0.001 \\
8 & 13.95  & < 0.001 \\
9  & -13.91 & < 0.001 \\
10 & 13.89  & < 0.001 \\
\hline
\end{tabular}
\end{table}

\subsubsection{Analysis of Key Differentiating Dimensions}
\begin{figure}[htbp]

        \centering
        \includegraphics[width=0.7\linewidth]{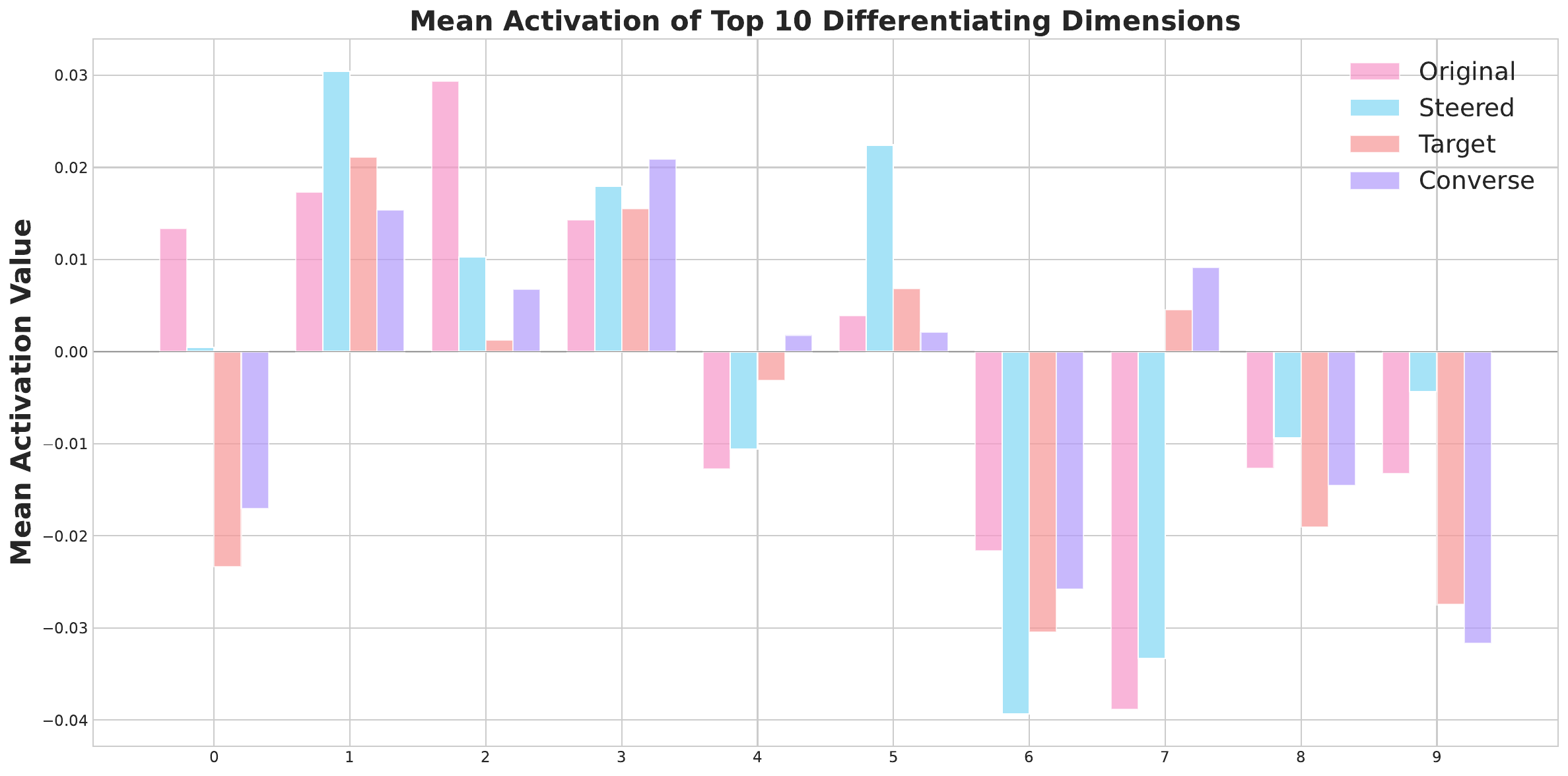}

    \caption{Analysis of steering across the Top 10 Differentiating Dimensions between target and converse embeddings. This figure portrays the effect of dimension-specific steering that the SteeringGate enables.}
    \label{top_dims}
\end{figure}

To investigate the specific semantic modifications induced by our steering vector, we analyzed the mean activation values across the top 10 differentiating dimensions between the target and converse prompts, as illustrated in Figure \ref{top_dims}. The plot provides compelling evidence for the efficacy of the steering method. A consistent pattern emerges where the steered embeddings have their activations systematically shifted away from the unsteered and converse embeddings, closer to the target embeddings. This validates that the steering successfully manipulates the core features that distinguish the target concept from its converse.
The overwhelming trend across the dimensions is one of successful and targeted semantic alignment. Therefore, this analysis validates that our steering mechanism operates by precisely modulating the key semantic features that define the target concept.

\newpage
\subsection{Qualitative Examples}
This section provides qualitative examples that visually demonstrate the effect and performance of our steering methodology.

\subsubsection{SteerVLM vs. Prompting}
 % For >{\bfseries}

% For demonstration purposes, let's assume the image is in the same directory
% or a subdirectory like 'figures/'
% \graphicspath{{figures/}} % Uncomment if your image is in a 'figures' subdirectory
This section presents a direct comparison between our SteerVLM method and a standard prompt-engineered approach for the same task. As shown with the example in Table \ref{tab:steered-vs-zero-image-left-aligned-nospace}, both methods were tasked with describing an image of a performer on stage. However, they were guided by opposing beliefs: the positive belief that "Learning new skills is exciting" and the negative belief that "Learning new skills is overwhelming." The resulting captions demonstrate how SteerVLM produces a nuanced description reflecting the "weight of the moment" and "a sense of pressure," in line with the negative belief, while the standard prompting method generates a more straightforward caption expressing "excitement or enthusiasm."

% \begin{document}

\begin{table*}[ht]
\centering % Centers the entire two-column layout on the page

% ---------- Left Column (image + explanatory text) -----------------
\begin{minipage}[c]{0.30\textwidth} % Adjust width as needed, 'c' for vertical centering
  \centering % Centers the image within this minipage
  \includegraphics[width=\linewidth]{figures/singing.png} % Image fills this minipage's width
  % \vspace{1em} % REMOVED: Space below the image was here

  { % Start a group for specific text formatting
    \small % Apply small font size to this text block
    \raggedright % Left-align text and remove paragraph indentation for this block
                 % \raggedright also sets \parindent to 0pt
    \textbf{Positive belief:} Learning new skills is overwhelming\par
    \textbf{Negative belief:} Learning new skills is exciting\par
    \vspace{0.4em} % Vertical space before the prompt text
    \textbf{Prompt:} Describe the image concisely.\par
  } % End group for text formatting
\end{minipage}\hfill % \hfill creates flexible space, pushing the next minipage to the right
% ---------- Right Column (comparison table) ----------------------------------------
\begin{minipage}[c]{0.68\textwidth} % Adjust width as needed
  \begin{tabularx}{\linewidth}{|>{\bfseries}l|X|} % Use \linewidth to fit table within this minipage
  \hline
  SteerVLM &
  The individual in the image appears to be on stage, possibly performing or engaging in an activity, wearing black pants and a black shirt. The background is dark, and the lighting is dim, suggesting that the setting might be a concert or a similar event. The person is holding a microphone, which could indicate they are singing or speaking. The expression on their face is uncertain, possibly conveying a range of \textbf{emotions, from focus to concentration. The overall atmosphere seems intense, with the person possibly feeling the weight of the moment.} The setting and the person's posture may evoke \textbf{a sense of pressure or a high stakes situation}, as they stand on stage. \\ \hline
  Prompting &
  The man on stage is wearing a black shirt and has his arms up in the air, possibly \textbf{expressing excitement or enthusiasm}. He appears to be the main focus of the scene, possibly indicating that he is the central figure in the performance. The stage appears to be dimly lit, which may create a more dramatic atmosphere for the performance. \\ \hline
  \end{tabularx}
\end{minipage}

\caption{SteerVLM vs.\ Prompt-engineered Zero-Shot captions produced for the same task and target and converse prompts}
\label{tab:steered-vs-zero-image-left-aligned-nospace}
\end{table*}

\subsubsection{Token-Level Analysis of Steering}
 This subsection provides a qualitative example that visualizes the steering effect at the individual token level. In the example below, the generated text is overlaid with a heatmap where the intensity of the color (darker blue) corresponds to the magnitude of the steering module's influence on the selection of that specific token.

\begin{figure}[htbp]

        \centering
        \includegraphics[width=0.8\linewidth]{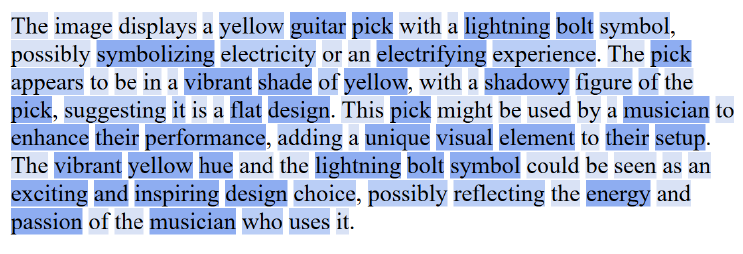}

    \caption{Figure depicting effect of steering for \textbf{Target prompt}: "Learning new skills is exciting" vs \textbf{Converse prompt}: "Learning new skills is overwhelming"}
    \label{token_level_1}
\end{figure}

\begin{figure}[htbp]

        \centering
        \includegraphics[width=0.8\linewidth]{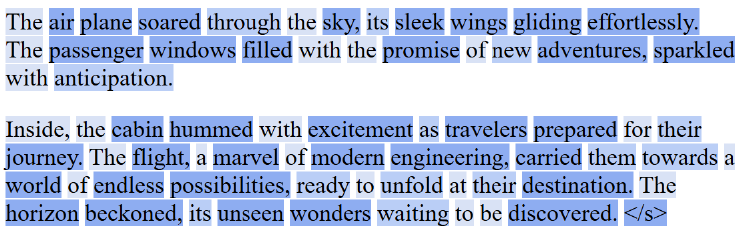}

    \caption{Figure depicting effect of steering for \textbf{Target prompt}: "Packing a vacation with diverse activities and constant exploration creates the most memorable travel experiences" vs \textbf{Converse prompt}: "True vacation rejuvenation comes from completely unwinding and relaxing at a single, peaceful destination"}
    \label{token_level_2}
\end{figure}

 In this example (Figure \ref{token_level_1}), the model describes a yellow guitar pick. We can observe that the steering influence is not uniform across the text. Instead, it is most pronounced on key descriptive words and thematic concepts that align with the intended steered output. For instance, tokens such as \textbf{electrifying, vibrant, exciting and inspiring, and energy and passion} show a significantly higher degree of steering. 
Similarly, in Figure \ref{token_level_2}, words such as \textbf{endless possibilities, unseen wonders to be discovered}, etc. show a higher degree of steering aligning to the target prompt.
 
 This demonstrates that the steering mechanism is not merely applying a general bias but is actively guiding the model to select specific, high-impact words that shape the narrative and sentiment of the description. This token-level attribution provides valuable insight into the precision and interpretability of our steering method.

\newpage
\subsubsection{Qualitative Comparison}
To situate our work within the context of existing research, we conducted a qualitative comparison against several other steering methods. In the example provided, an image of a floral arrangement is presented alongside a target prompt ("Filling your home with many plants makes it feel alive and truly welcoming") and a converse prompt ("Keeping just a few simple plants is key to a clean and tidy living space"). The subsequent Table \ref{tab:comparison} showcases the image descriptions generated by various models, including our own. This comparison highlights the different ways each method interprets the image to align with the given prompt, offering a clear view of our method's ability to create a harmonious and contextually rich narrative.

\begin{center} % To center the minipage containing the image and text
\begin{minipage}{0.9\textwidth} % Adjust width as needed for this top block
  \centering % Center content within this minipage
  
  % --- Image ---
  % REDUCE THE IMAGE WIDTH HERE:
  \includegraphics[width=0.4\linewidth]{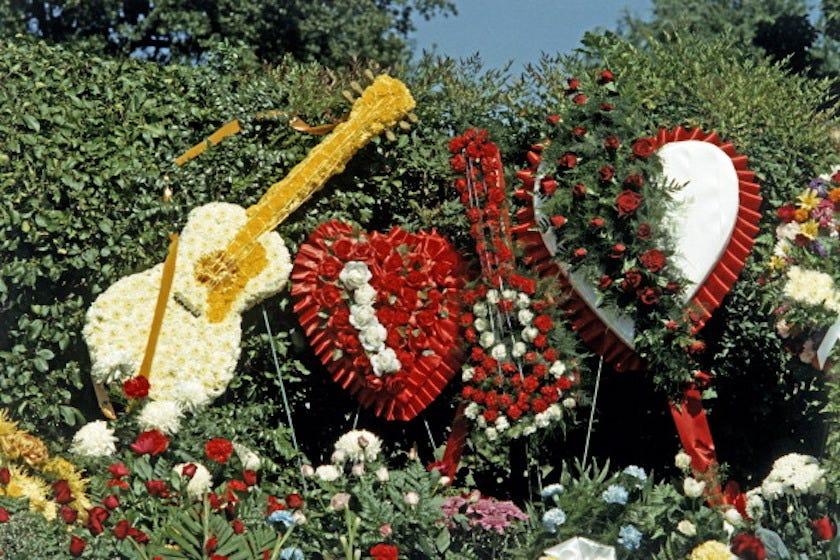} % <<<< TRY 0.4\linewidth or 0.35\linewidth
  % Or use an explicit height if the aspect ratio is very tall:
  % \includegraphics[height=5cm]{figures/comparison.png} 

  % --- Explanatory Text ---
  { % Start a group for specific text formatting
    \raggedright 
    \textbf{Target Prompt:} Filling your home with many plants makes it feel alive and truly welcoming.\par
    \vspace{0.4em} 
    \textbf{Converse Prompt:} Keeping just a few simple plants is key to a clean and tidy living space.\par
    \vspace{0.4em} 
    \textbf{Prompt:} Describe the following image in detail. Ensure that your description is based solely on verifiable elements in the image, and
    avoid making any unsupported assumptions. If uncertainty is necessary, ensure that you include qualifiers such as 'maybe', 'appearing to', or
    'possibly'. Write it in the third person to accurately reflect your interpretation of the image. \par
  } 
\end{minipage}
\end{center}

% REDUCE THIS SPACE IF NEEDED:
\vspace{1em} % <<<< TRY 1em or 0.8em instead of 1.5em

% ---------- Combined Table for 6 Samples (using ltablex) --------------------
% \keepXColumns % Make sure this is active
\newpage
\begin{table}[!ht]

% --- Caption and Label ---
% Place these inside the 'table' environment, BEFORE the 'tabularx' environment.
\caption{Qualitative comparison of existing steering methods on an example from the \textsc{Vnia} evaluation dataset.}
\label{tab:comparison}

% --- The Table Itself ---
% Set font size for the whole table here. This is cleaner than adding \small to every row.
\small 
% Optional: Increase row spacing for better readability
\renewcommand{\arraystretch}{1.3} 

\begin{tabularx}{\linewidth}{|>{\bfseries}l|X|}

% --- Table Header ---
% The header is now the first row of the table, defined in a standard way.
\hline
\textbf{Method} & \textbf{Description} \\ 
\hline

% ----- Table Content -----
% (Your table rows here...)
\small Contrastive / layer &
The image features a beautiful flower arrangement in the shape of a heart, made with red and white flowers. The heart is adorned with a guitar, which adds a unique touch to the floral display. The guitar is positioned towards the left side of the heart, and the flowers are arranged around it, creating a visually appealing and creative display. \\ \hline
\small Act-Add \cite{turner2023actadd} &
The image showcases a beautiful display of flowers and decorations arranged in the shape of a heart. The heart is prominently displayed, surrounded by a plethora of flowers in various colors. The heart appears to be made of a combination of flowers and a guitar-shaped structure.

A teddy bear is also present in the scene, placed near the heart, adding to the overall charm of the display. The heart, flowers, and teddy bear together create a visually appealing arrangement that captures the essence of love and warmth. \\ \hline
\small ML-ACT \cite{rodriguez2024controllinglanguagediffusionmodels} &
The image features a beautiful arrangement of flowers and a guitar-shaped heart. The heart is adorned with a guitar, and it is surrounded by a variety of flowers. The flowers are arranged in a way that creates a visually appealing display. The heart is positioned in the center of the scene, drawing attention to its unique shape and design. The flowers surrounding the heart add a sense of depth and color to the overall composition of the image. \\ \hline
CAA \cite{marks2024steering} &
The image features a beautiful display of flowers and a guitar, creating a heartwarming scene. The guitar is prominently placed in the center of the scene, surrounded by a variety of flowers. The flowers are arranged in a way that showcases their vibrant colors and creates a visually appealing display. The guitar's position and the abundance of flowers suggest that this arrangement might be a tribute or a celebration of music and art. \\ \hline
\small MLLM Steering \cite{zhao2024analyzing} & <blank> \\
\hline

\small ACT \cite{ding2024adaptive}&
The image features a beautiful arrangement of flowers and a guitar. The guitar is placed in the center of the scene, surrounded by a variety of flowers. The flowers are scattered throughout the image, with some located near the guitar and others positioned further away. The combination of the guitar and the flowers creates a visually appealing scene. \\ \hline
\small Ours &
This image presents a vibrant display of a heart-shaped arrangement, adorned with a guitar and a violin, all set against a backdrop of lush greenery. The heart, likely a focal point, appears to be a symbol of love and affection. The violin and guitar, both stringed instruments, add a touch of music and creativity to the scene. The surrounding greenery, possibly a garden or a park, enhances the sense of tranquility and warmth. The combination of these elements creates a visually pleasing and inviting atmosphere, suggesting a space where one can enjoy the beauty of nature and the joy of music. The image captures a moment of harmony, where the elements come together to create a picturesque and welcoming environment. \\ \hline
\end{tabularx}
\end{table}

\newpage
\subsubsection{Analysis of Robustness to Semantic Variations in Prompt Pairs} \label{semantic_shift_analysis}

\paragraph{Prompt to generate natural language variation pairs}

\begin{table}[htbp]
\centering
\caption{Full Prompt for Generating Target-Converse Pairs with Natural Language Variation}
\label{tab:prompt_semantic_shift}
\begin{tabularx}{\textwidth}{p{0.9\textwidth}} % Using the X column is best practice for tabularx
    \toprule
    \textbf{Prompt} \\
    \midrule
    \begin{minipage}[t]{\linewidth} % This minipage isolates the content, preventing errors
    Given the following anchor prompts: Positive framing (p+): "\{target anchor prompt\}" Negative framing (p-): "\{converse anchor prompt\}"

    Generate 5 new prompt pairs that:
    \begin{enumerate}
        \item Have medium to high semantic similarity to the anchors. Preserve the positive (p+) or negative (p-) framing.
        \item Express the same core activity using different wordings, tones, or contexts.
        \item Include some pairs that mix anchor prompts (e.g., p+ vs anchor p-, anchor p+ vs anchor p-).
        \item Include one neutral p+ and one neutral p- mixing it with the anchor prompt.
    \end{enumerate}
    Use simple sentences. Return results clearly labeled as (p+) and (p-) pairs.
    \end{minipage} \\ % The \\ is still needed to end the table row
    \bottomrule
\end{tabularx}
\end{table}

\paragraph{Results}

To rigorously evaluate the robustness of SteerVLM to linguistic variations, we conducted an experiment analyzing its performance when subjected to semantic shifts in prompt pairings. We selected 3 prompt pairs from the 5 steering vectors extracted (as defined in Section \ref{evaluation_setup}) as baseline "anchor" prompt pairs. For each anchor, we then generated several "Target" and "Converse" variations, ensuring they maintained medium-to-high semantic similarity using GPT-4o \cite{openai2024gpt4ocard} as defined in Table \ref{tab:prompt_semantic_shift}. Semantic similarity was quantified using the cosine similarity of sentence embeddings from the Qwen3-8B model \cite{yang2025qwen3technicalreport}. SteerVLM's performance for each pair was measured using the same evaluation metrics set up in Section \ref{experiments}.
The results, presented in Table \ref{tab:semantic_shift}, reveal a consistent pattern across all three domains. SteerVLM’s performance remains remarkably stable, staying within a functional range of the anchor performance even when prompt similarity scores vary. The most notable performance differences occurred when prompt variations introduced more abstract or tangentially related concepts (e.g., shifting from the sensory `energizing/overwhelming' pair to the more cognitive `awake/chaotic' pair). This pattern suggests that the model is highly robust to direct paraphrasing and slight conceptual shifts, underscoring its ability to generalize based on core semantic intent rather than being reliant on specific keyword matching.

\begin{table}[]
\caption{SteerVLM Performance Analysis under Semantic Prompt Variations. The table presents results from three experiments evaluating model robustness to linguistic shifts in prompt pairings. For each domain, an "anchor" pair was established as a baseline. The "Target" and "Converse" columns show the varied prompt pairs used in testing. }
\label{tab:semantic_shift}
\small
\begin{tabular}{p{4.5cm}|p{4.5cm}|p{2cm}|p{2cm}|l}
\textbf{Anchor Target Prompt}                                                   & \textbf{Anchor Converse Prompt}                                                & \textbf{Sim target (prompt, anchor)} & \textbf{Sim converse (prompt, anchor)} & \textbf{Score} \\  \midrule \midrule
"Volunteering feels   fulfilling"                                               & "Volunteering feels like an obligation"                                        &                                             &                                        & 0.63                 \\
                                                                                &                                                                                &                                             &                                        &                      \\
\textbf{Target Prompt}                                                          & \textbf{Converse Prompt}                                                       &                                             &                                        &                      \\ \midrule
"Helping others makes me   feel proud"                                          & "Helping others feels like a duty"                                             & 0.764                                       & 0.83                                   & 0.54                 \\
"Volunteering gives me a   sense of purpose"                                    & "Volunteering feels stressful and forced"                                      & 0.89                                        & 0.88                                   & 0.61                 \\
"Lending a hand brings   joy"                                                   & "Volunteering feels like an obligation"                                        & 0.81                                        & 1                                      & 0.72                 \\
"Volunteering feels   fulfilling"                                               & "Sometimes helping others feels tiresome"                                      & 1                                           & 0.63                                   & 0.63                 \\
"Occasionally helping   out can be rewarding"                                   & "Sometimes volunteering feels like a chore"                                    & 0.734                                       & 0.784                                  & 0.63                 \\
                                                                                &                                                                                &                                             &                                        &                      \\
\textbf{Anchor Target Prompt}                                                   & \textbf{Anchor Converse Prompt}                                                & \textbf{Sim target (prompt, anchor)} & \textbf{Sim converse (prompt, anchor)} & \textbf{Score} \\ \midrule \midrule
"Bright   colors are energizing"                                                & "Bright colors are overwhelming"                                               &                                             &                                        & 0.84                 \\
                                                                                &                                                                                &                                             &                                        &                      \\
\textbf{Target Prompt}                                                          & \textbf{Converse Prompt}                                                       &                                             &                                        &                      \\ \midrule
"Vibrant colors lift my   mood"                                                 & "Vibrant colors feel too intense"                                              & 0.867                                       & 0.92                                   & 0.865                \\
"Bright shades make me   feel awake"                                            & "Bright shades feel chaotic"                                                   & 0.834                                       & 0.85                                   & 0.67                 \\
"Colorfulness makes me   cheerful"                                              & "Bright colors are overwhelming"                                               & 0.79                                        & 1                                      & 0.825                \\
"Bright colors are   energizing"                                                & "Some colors feel too strong"                                                  & 1                                           & 0.87                                   & 0.835                \\
"Bold tones are   nice"                                                         & "Bold tones create tension"                                                    & 0.77                                        & 0.75                                   & 0.835                \\
                                                                                &                                                                                &                                             &                                        &                      \\
\textbf{Anchor Target Prompt}                                                   & \textbf{Anchor Converse Prompt}                                                & \textbf{Sim target (prompt, anchor)} & \textbf{Sim converse (prompt, anchor)} & \textbf{Score} \\ \midrule \midrule
"Filling your home with   many plants makes it feel alive and truly welcoming." & "Keeping just a few simple plants is key to a clean and   tidy living space.", &                                             &                                        & 0.69                 \\
                                                                                &                                                                                &                                             &                                        &                      \\
\textbf{Target Prompt}                                                          & \textbf{Converse Prompt}                                                       &                                             &                                        &                      \\ \midrule
"A house full of   greenery feels warm and inviting."                           & "Too many plants make a home feel messy and hard to   manage."                 & 0.8                                         & 0.75                                   & 0.79                 \\
"Adding some plants can   give the room a gentle touch of life."                & "Keeping just a few simple plants is key to a clean and   tidy living space.", & 0.81                                        & 1                                      & 0.705                \\
"Filling your home with   many plants makes it feel alive and truly welcoming." & "A room with fewer plants feels simpler and easier to   care for.",            & 1                                           & 0.83                                   & 0.715                \\
"Caring for lots of   plants makes a space feel loved and full of energy."      & "Limiting plants to one or two keeps things minimal and   stress-free.",       & 0.84                                        & 0.8                                    & 0.67                 \\
"Surrounding yourself   with many plants creates a refreshing escape indoors."  & "Choosing almost no plants keeps the space feeling open   and uncluttered."    & 0.83                                        & 0.77                                   & 0.685               
\end{tabular}
\end{table}

 \newpage
\subsection{Steering Module Architecture Block Diagram}

% write about module choices here

\begin{figure}[htbp]
    \centering
    \begin{subfigure}[t]{0.48\linewidth}
        \centering
        \includegraphics[width=0.6\linewidth]{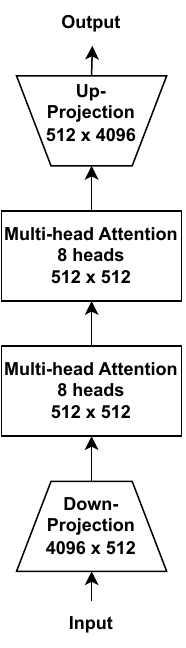}
        \caption{Steerer}
        \label{steerer}
    \end{subfigure}
    \hfill
    \begin{subfigure}[t]{0.48\linewidth}
        \centering
        \includegraphics[width=0.6\linewidth]{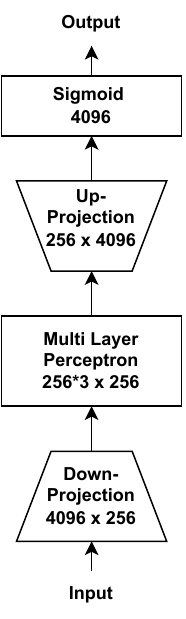}
        \caption{Steering Gate}
        \label{steeringgate}
    \end{subfigure}
    \caption{Steering Module Block Diagram specifically for the LLaVA1.5-7B model}
    \label{steering_module_arch}
\end{figure}

% Insert block diagram of the Steerer here.
Figure \ref{steering_module_arch} provides a schematic of the steering module architecture specifically designed for integration with the LLaVA1.5-7B model. 

During the experimental phase, we evaluated a variety of architectural choices for both the Steerer and the SteeringGate. Our initial attempts employed MLP-based architectures for both modules. While functional, these configurations were computationally expensive and did not yield meaningful performance improvements. Similarly, incorporating attention layers into the SteeringGate led to reduced qualitative performance.

We tested variants with 1, 2, and 3 attention layers, as well as 1- and 2-layer MLPs. A single attention layer resulted in unstable training, with signs of early overfitting and poor qualitative outcomes. Increasing the depth to 2 or 3 layers provided similar qualitative results, but at a higher computational cost, so we opted for the more parameter-efficient design. For the SteeringGate, a 1-layer MLP provided a favorable balance of training stability, computational efficiency, and qualitative performance.

Because attention layers scale quadratically with dimension, we also explored downprojection strategies to reduce parameter count. Specifically, we compared projection dimensions of 512 and 256. For the Steerer, a dimensionality of 256 yielded too few parameters, which led to unstable training and higher evaluation loss. For the SteeringGate, however, the performance difference between 512 and 256 dimensions for the MLP was qualitatively negligible, hence we selected 256 as the more efficient option.

All qualitative evaluations for the architectures were conducted on a subset of the VNIA evaluation dataset.

The diagram is divided into two main components: the Steerer (a) and the Steering Gate (b). The Steerer processes the input through a series of down-projection, 2-layer multi-head attention, and up-projection layers to generate the steering influence. The Steering Gate utilizes a multi-layer perceptron (MLP) and a sigmoid activation function to control the flow and intensity of the steering signal. This modular design allows for effective and controlled guidance of the vision language model's output.

\end{document}